%% file: main.tex
\pdfoutput=1
\documentclass[11pt, a4paper, logo]{gdm/googledeepmind}


\pdftrailerid{redacted}

\makeatletter
\renewcommand\bibentry[1]{\nocite{#1}{\frenchspacing\@nameuse{BR@r@#1\@extra@b@citeb}}}
\makeatother

\usepackage{lipsum}
\usepackage{dsfont}
\usepackage{gdm/gdm-colors}

\usepackage[authoryear, sort&compress, round]{natbib}
\usepackage{comment}
\graphicspath{{figures/}}

\usepackage{lineno}
\input{contents/math_commands}

\usepackage[colorlinks=true,allcolors=blue]{hyperref}
\hypersetup{
citecolor=[HTML]{2420cf}
}

\hypersetup{
linkcolor=[HTML]{2420cf}
}
\usepackage{url}

\usepackage[utf8]{inputenc} %
\usepackage[T1]{fontenc}    %
\usepackage{url}            %
\usepackage{booktabs}       %
\usepackage{amsfonts}       %
\usepackage{nicefrac}       %
\usepackage{microtype}      %
\usepackage{xcolor}         %

\usepackage{bbm}
\usepackage{pifont}

\usepackage{algpseudocode}
\usepackage[ruled,vlined,linesnumbered]{algorithm2e}

\usepackage{microtype}
\usepackage{graphicx}
\usepackage{booktabs} %
\usepackage{colortbl}

\usepackage{algorithm2e}

\usepackage{amsfonts}   %
\usepackage[table]{xcolor}
\usepackage[most]{tcolorbox}

\usepackage{enumitem}

\usepackage{amssymb}
\usepackage{mathtools}
\usepackage{amsthm}
\usepackage{amsmath}
\usepackage{multirow}
\usepackage{makecell}
\usepackage{appendix}
\usepackage{enumitem}
\usepackage{nccmath}
\usepackage{wrapfig}
\definecolor{darkgreen}{rgb}{0.0, 0.5, 0.0}
\usepackage[capitalize,noabbrev]{cleveref}

\newtcolorbox{compactgraybox}{
  colback=black!4,
  colframe=black,
  boxrule=0.4pt,
  arc=2.4pt,
  left=1.2mm,
  right=1.2mm,
  top=1.2mm,
  bottom=1.2mm,
  boxsep=0.6mm,
  before skip=0.4em,
  after skip=0.4em
}
\definecolor{MintGreen}{RGB}{175,240,175}

\NewDocumentCommand{\lyx}
{ mO{} }{\textcolor{orange}{\textsuperscript{\textit{lyx}}\textsf{\textbf{\small[#1]}}}}

\NewDocumentCommand{\hongru}
{ mO{} }{\textcolor{blue}{\textsuperscript{\textit{Hongru}}\textsf{\textbf{\small[#1]}}}}

\NewDocumentCommand{\cheng}
{ mO{} }{\textcolor{orange}{\textsuperscript{\textit{Cheng}}\textsf{\textbf{\small[#1]}}}}

\NewDocumentCommand{\Jiahao}
{ mO{} }{\textcolor{green}{\textsuperscript{\textit{Jiahao}}\textsf{\textbf{\small[#1]}}}}
\usepackage{siunitx}
\usepackage{tikz}
\usepackage{xcolor}
\usepackage{graphicx}
\usetikzlibrary{positioning, fit, calc}
\definecolor{lightpink}{RGB}{255,230,230}
\definecolor{lightblue}{RGB}{230,240,255}
\definecolor{highlightpink}{RGB}{255,180,180}
\definecolor{highlightblue}{RGB}{180,210,255}
\definecolor{highlightgreen}{RGB}{180,255,180}
\definecolor{LavenderLight}{HTML}{C7C3F5}

\renewcommand{\eqref}[1]{Eq.~\ref{#1}}

\theoremstyle{plain}

\theoremstyle{definition}

\theoremstyle{remark}

\usepackage{marvosym}
\usepackage{caption}
\usepackage[most]{tcolorbox}
\usepackage{amssymb}
\usepackage{bm}

\definecolor{impr}{RGB}{34, 139, 34}
\definecolor{lightred}{RGB}{255, 230, 230}
\definecolor{darkred}{RGB}{192, 0, 0}

\definecolor{bestbg}{HTML}{FFF2B2}   %
\definecolor{secondbg}{HTML}{E8F0FE} %
\definecolor{LavenderLight}{HTML}{C7C3F5}
\definecolor{LightCoral}{RGB}{240,128,128}
\definecolor{LightBlue}{RGB}{173,216,230}

\tcolorboxenvironment{remark}{
    colback=LightCoral!10,       %
    colframe=black,     %
    arc=1mm,               %
    top=.6mm,
    bottom=.6mm,
    boxrule=.5pt,           %
    right=1mm,
    fonttitle=\bfseries,   %
    left=1mm,              %
    attach boxed title to top left={yshift=-2mm, xshift=4mm},
    boxed title style={
        colback=black!40,
        arc=1mm,
    },
    before skip=6pt,   %
    after skip=6pt,    %
}

\tcolorboxenvironment{theorem}{
    colback=LightBlue!20,       %
    colframe=black,     %
    arc=1mm,               %
    top=.6mm,
    bottom=.6mm,
    right=1mm,
    boxrule=.5pt,           %
    fonttitle=\bfseries,   %
    left=1mm,              %
    attach boxed title to top left={yshift=-2mm, xshift=4mm},
    boxed title style={
        colback=black!40,
        arc=1mm,
    },
    before skip=6pt,   %
    after skip=6pt,    %
}

\tcolorboxenvironment{lemma}{
    colback=red!10,       %
    colframe=black!,     %
    arc=1mm,               %
    boxrule=1pt,           %
    fonttitle=\bfseries,   %
    left=1mm,              %
    attach boxed title to top left={yshift=-2mm, xshift=4mm},
    boxed title style={
        colback=black!40,
        arc=1mm,
    }
}

\title{\textit{From Word to World}: Can Large Language Models be Implicit Text-based World Models?}

\renewcommand{\today}{}

\author[1,2]{Yixia Li}
\author[3, \Letter]{Hongru Wang}
\author[4]{Jiahao Qiu}
\author[5]{Zhenfei Yin}
\author[2]{Dongdong Zhang}
\author[6]{Cheng Qian}
\author[7]{Zeping Li}
\author[8]{Pony Ma}
\author[1, \Letter]{\mbox{Guanhua Chen}}
\author[6]{Heng Ji}

\affil[1]{Southern University of Science and Technology}
\affil[2]{Microsoft Research}
\affil[3]{University of Edinburgh}
\affil[4]{Princeton University}
\affil[5]{Oxford University}
\affil[6]{\mbox{University of Illinois Urbana-Champaign}}
\affil[7]{Fudan University}
\affil[8]{Mind Lab}
\affil[\Letter]{Corresponding Authors }

\input{contents/0_abstract.tex}

\begin{document}

\maketitle

\addtocontents{toc}{\protect\setcounter{tocdepth}{-1}}
\input{contents/1_introduction.tex}

\input{contents/2_relatedwork.tex}
\input{contents/3_llm_as_world_models.tex}
\input{contents/4_evaluation.tex}
\input{contents/5_fidelity.tex}
\input{contents/6_scalability.tex}
\input{contents/7_agent_utility.tex}

\input{contents/8_conclusion.tex}

\bibliographystyle{abbrvnat}
\nobibliography*
\bibliography{main}

\input{contents/appendix.tex}

\end{document}

%% file: contents/math_commands.tex
\usepackage{amsmath,amsfonts,bm}

\def\eqref#1{equation~\ref{#1}}

\def\1{\bm{1}}

\DeclareMathAlphabet{\mathsfit}{\encodingdefault}{\sfdefault}{m}{sl}
\SetMathAlphabet{\mathsfit}{bold}{\encodingdefault}{\sfdefault}{bx}{n}

\usepackage{amsthm}

\newcommand\TODO[1][]{{\color{orange}[TODO\ifthenelse{\equal{#1}{}}{}{: #1}]}}
\newcommand\SEC\section
\newcommand\SSEC\subsection
\newcommand\SSSEC\subsubsection

%% file: contents/0_abstract.tex
\begin{abstract}
\vspace{-18pt}
{\fontsize{10pt}{10pt} \selectfont \raisebox{-0.1em}{\includegraphics[height=1em]{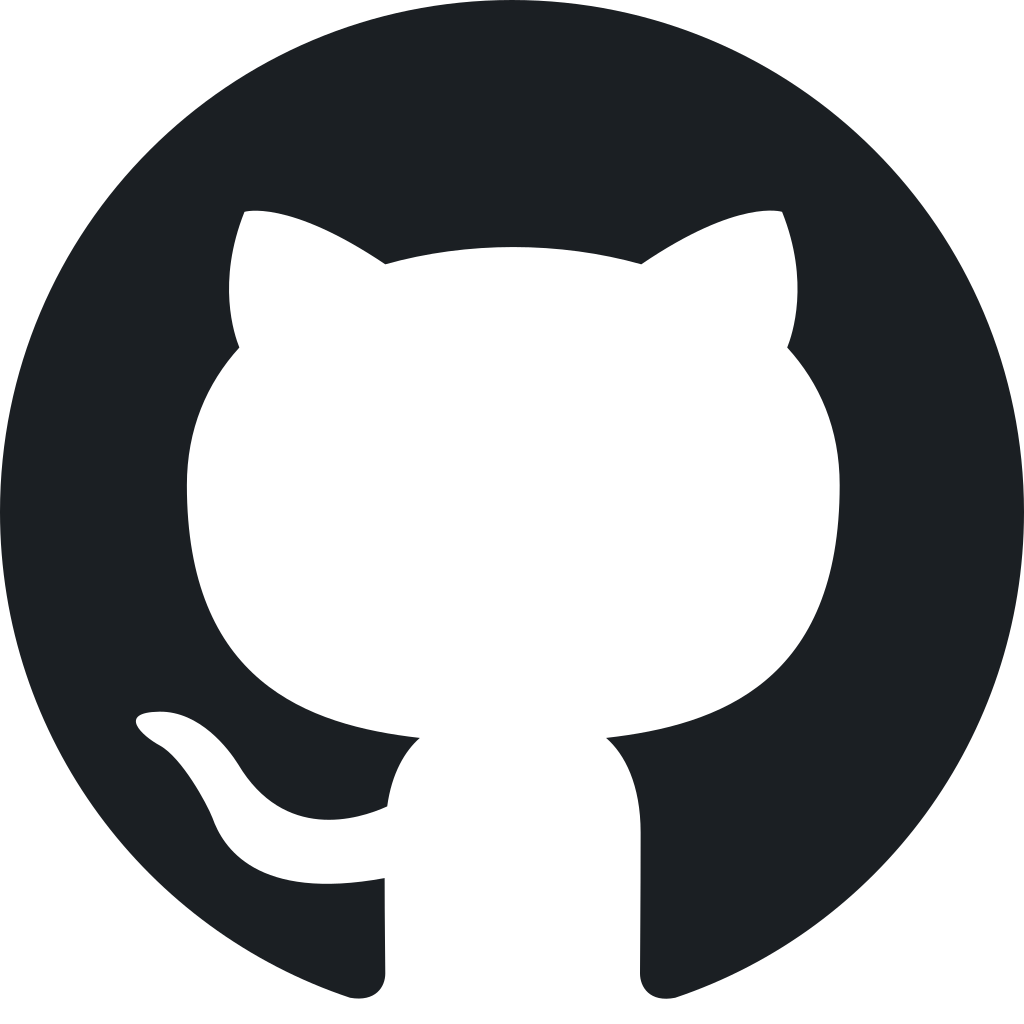}} Code: \href{https://github.com/X1AOX1A/Word2World}{https://github.com/X1AOX1A/Word2World}}
\\[0.6em]
    Agentic reinforcement learning increasingly relies on experience-driven scaling, yet real-world environments remain non-adaptive, limited in coverage, and difficult to scale. World models offer a potential way to improve learning efficiency through simulated experience, but it remains unclear whether large language models can reliably serve this role and under what conditions they meaningfully benefit agents. We study these questions in text-based environments, which provide a controlled setting to reinterpret language modeling as next-state prediction under interaction. We introduce a three-level framework for evaluating LLM-based world models: (i) fidelity and consistency, (ii) scalability and robustness, and (iii) agent utility. Across five representative environments, we find that sufficiently trained world models maintain coherent latent state, scale predictably with data and model size, and improve agent performance via action verification, synthetic trajectory generation, and warm-starting reinforcement learning. Meanwhile, these gains depend critically on behavioral coverage and environment complexity, delineating clear boundry on when world modeling effectively supports agent learning.
\end{abstract}

%% file: contents/1_introduction.tex
\begin{figure*}[h]
    \centering
    \includegraphics[width=1\textwidth]{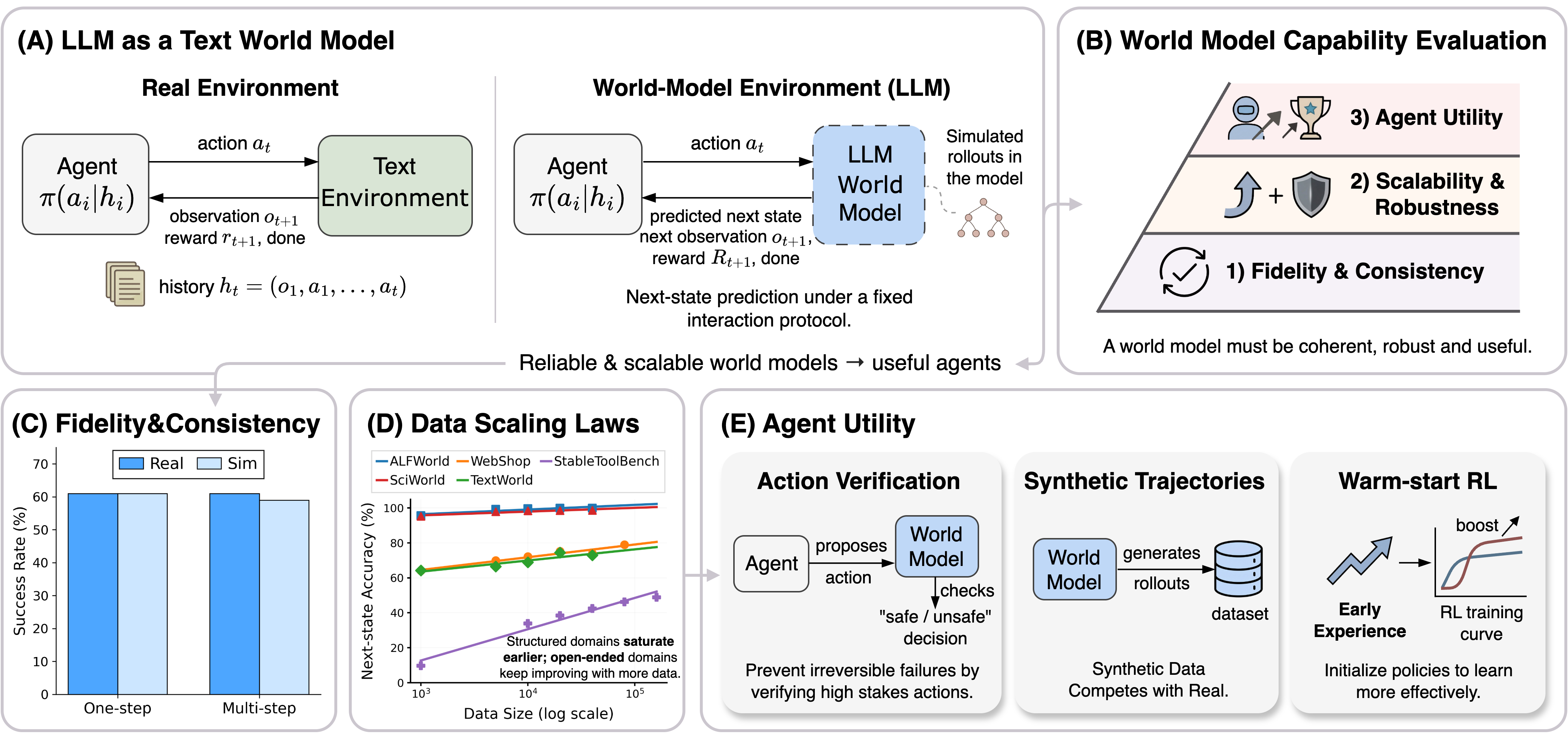}
    \caption{
\textbf{LLMs as text-based world models for agent learning.}
(A) We formulate world modeling as next-state prediction under a fixed text-based interaction protocol.
(B) Assess world-model capability along three axes: fidelity/consistency, scalability/robustness, and agent utility.
(C) World model exhibits high fidelity and consistency in both single-step predictions and long-horizon rollouts.
(D) Performance scales predictably with increased training data across text environments.
(E) Faithful world models enhance agents via verification, synthetic data generation, and improved reinforcement learning through stronger initialization.
    }
    \label{fig:main}
\end{figure*}

\section{Introduction}

Recent progress in agentic reinforcement learning (RL) increasingly hinges on \textbf{experience-driven scaling}: as agents improve, further gains demand larger, more diverse, and more challenging environments \citep{zeng2025rlvescalingreinforcementlearning,zhang2025autoenvautomatedenvironmentsmeasuring,tong2025gamerlsynthesizingmultimodalverifiable}. Unlike static pretraining corpora, however, experience must be collected through interaction. As agents scale, this reliance exposes a fundamental \textbf{experience bottleneck}: realistic environments are non-adaptive, difficult to scale, and limited in coverage, which become key limiting factors for progress in agentic RL \citep{wei2025webagentr1trainingwebagents,jiang2025verltoolholisticagenticreinforcement,guo2025stabletoolbenchstablelargescalebenchmarking}.

A key lever for alleviating this bottleneck is world modeling \citep{hafner2024masteringdiversedomainsworld,hafner2025trainingagentsinsidescalable,zhao2025generalist,hu2025textworld}. Effective learning from interaction requires agents to maintain internal models of the environment that track latent state, predict action outcomes, and learning beyond immediate observations. By internalizing environment dynamics and enabling learning from imagined interaction, world models help close the interaction loop and enable more efficient and effective learning. Meanwhile, LLM trained at massive scale via next-token prediction, exhibit strong generalization and encode rich world knowledge \citep{grattafiori2024llama3herdmodels,qwen2025qwen25technicalreport,hu2025unveiling}. This naturally raises the question:

\begin{center}
\vspace{0.1em}
\includegraphics[height=1em]{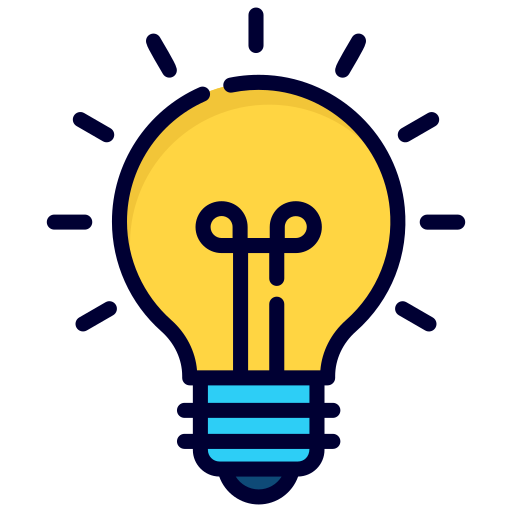}\hspace{0.3em}%
{\itshape\bfseries Can large language models serve as effective world models, \\thereby improving agents to learn from experience?}
\vspace{0.1em}
\end{center}

While prior work has explored LLMs as simulators, experience generators, or planning interfaces \citep{chen2025scalingagentlearningexperience,li2025simulatingenvironmentsreasoningmodels,wu2025rlvrworldtrainingworldmodels,gu2025llmsecretlyworldmodel,wang2025world,he2025pretrained}, it remains unclear \emph{how} to learn a world model and \emph{when} it is reliable enough to improve downstream agents. A useful world model must go beyond locally plausible text, maintaining coherent state over time, remaining robust to distribution shift, and providing measurable utility. To address these questions in a controlled setting, we focus on \textbf{text-based environments} as a unifying interface between language modeling and world modeling. This abstraction preserves core challenges of agentic-environment interaction while reframing the objective from \textit{next-token prediction} to \textit{next-state prediction} under a fixed interaction protocol.

Through this lens, we formalize a three-level framework for characterizing world modeling capabilities in agent learning.

\begin{compactgraybox}
    \begin{itemize}[leftmargin=1.3em, topsep=0.2em, itemsep=0.25em, parsep=0pt, partopsep=0pt]
        \item \textbf{Fidelity and Consistency:} Examine whether a world model maintains coherent latent states over short and long horizons, forming the foundation for reliable world modeling.
        \item \textbf{Scalability and Robustness:} Characterize how world modeling capabilities scale and generalize under varying conditions, providing practical guidance for world model development.
        \item \textbf{Agent Utility:} Evaluate whether high-fidelity world models translate into measurable improvements for downstream agents.
    \end{itemize}
\end{compactgraybox}

Studying five representative text-based environments, our analysis yields three main findings:
(i) LLMs can function as reliable world models: they exhibit internal latent dynamics that support in-context world modeling, and supervised fine-tuning substantially improves short-term predictive fidelity and enables consistent long-horizon rollouts in well-structured domains.
(ii) The development of reliable world models requires systematic scaling of model capacity and data volume with environment complexity, and maintaining robustness to distribution shift through broad behavioral coverage and diverse environment exposure.
(iii) Fidelity world models provide practical benefits to agents by enabling verification of high-stakes actions to prevent irreversible failures, and by generating synthetic trajectories and warm-starting reinforcement learning to improve learning efficiency and effectiveness.

Taken together, these findings illuminate both the promise and the limits of LLM-based world models in text environments. From words to worlds, from next-token to next-state prediction, we provide an empirical foundation for treating LLMs as general-purpose world models for agentic learning and chart a path toward domains beyond text.

%% file: contents/2_relatedwork.tex
\section{Related Works}

Large language models have recently been explored as world models across a variety of text-based and structured settings. Prior efforts in world modeling largely focus on predicting environment dynamics through structured or discrete state representations. Patch-based approaches prompt LLMs to estimate state deltas in ByteSized32 \citep{wang2024language,yang2024evaluatingworldmodelsllm}, while in web navigation, systems such as WMA \citep{chae2025webagentsworldmodels} and RLVR-World \citep{wu2025rlvrworldtrainingworldmodels} reason over updates to the Accessibility Tree. Other lines of work adopt closed-form prediction schemes where the model outputs predefined symbolic labels or categories, including preconditions and effects in cooking environments \citep{xie2024making}, disaster impact ratings \citep{li2025llmsworldmodelsdatadriven}, or classifier-head predictions trained on LLM embeddings \citep{yang-etal-2025-efficient}. Although these methods illustrate the utility of structured prediction for specific settings, they generally depend on environment-specific abstractions and a fixed output space tailored to particular domains. In contrast, we formulate world modeling as a multi-turn natural language simulation task, where the LLM generates next-state transitions in free text, enabling more general and compositional interaction patterns.

Regarding model adaptation, much prior work employs zero-shot or few-shot prompting \citep{wang2024language,yang2024evaluatingworldmodelsllm,li2025llmsworldmodelsdatadriven,zuo2025radi} or attaches lightweight classifier heads for closed-form prediction \citep{yang-etal-2025-efficient}. While such settings highlight the latent capabilities of LLMs, they often yield limited accuracy and constrains their applicability in downstream tasks.
Moving beyond prompting-based adaptations, we finetune LLMs on large-scale multi-turn interaction trajectories to better internalize environment dynamics over extended horizons.

Prior evaluation efforts largely center on single-step prediction accuracy in limited environments and domains \citep{wang2024language,xie2024making,chae2025webagentsworldmodels,li2025llmsworldmodelsdatadriven} , and rarely examines long-horizon consistency or compounding errors—factors that are critical for using world models as reliable simulators. Consequently, it remains an open question whether LLM-based world models can produce coherent multi-step trajectories that are executable in real environments. To address this gap, we conduct a systematic evaluation across five representative environments, measuring not only one-step fidelity but also rollout stability, WM-to-Real transfer, and generalization across agents, environments, and scales.

%% file: contents/3_llm_as_world_models.tex
\section{LLM as Text-based World Models}
\subsection{Formalization of World Models}

We formalize the interaction between an \textbf{agent} and a \textbf{text-based world model} as a multi-turn language-based decision process, where both perception and action are represented in natural language.

\paragraph{Agent}
A text-based agent $\mathcal{A}$ operates in a ReAct style \citep{yao2023reactsynergizingreasoningacting}, yielding a simple, unified interface where each step involves internal reasoning and external action.
Formally, the agent is defined as:
\begin{equation}
\mathcal{A}: \{ S_0, (T_i, A_i, S_i)_{i=1}^{n-1}\} \rightarrow (T_n, A_n),
\label{eq:agent}
\end{equation}
where $S_i$ denotes the textual observation (or environment response) at step $i$,
$T_i$ represents the agent’s internal reasoning trace,
and $A_i$ denotes the explicit action expressed in natural language.

\paragraph{World Model}
The environment or its surrogate world model $\mathcal{W}$ defines the complementary mapping:

\begin{equation}
\mathcal{W}: \{ S_0, (A_i, S_i')_{i=1}^{n-1}, A_n\} \rightarrow (S_n', R_n'),
\label{eq:worldmodel}
\end{equation}
where $S_n'$ denotes the next state predicted by the world model, and $R_n' \in \{0, 1\}$ is a binary reward indicating task success or termination.
A value of $R_n' = 1$ corresponds to a successful completion, while $R_n' = 0$ denotes either an unfinished or failure state (e.g., triggering validation at the wrong time).
Through these textual transitions, the world model functions as an implicit \textbf{next-state predictor} of environment dynamics.
This capability can be realized through \textbf{in-context learning}, where the model leverages few-shot examples of state transitions in its prompt, or through \textbf{supervised fine-tuning} on trajectory data to learn the underlying dynamics.

Note that in practice, text-based environments are inherently POMDPs (Partially Observable Markov Decision Processes): the true environment state is richer than what is described to the agent in text. For example, in ALFWorld a room may contain objects and spatial details that are never mentioned (e.g., what is inside a closed drawer), yet these hidden factors matter for predicting how the world evolves.
Thus, although the agent only receives a partial view of the initial state $S_0$, the world model can be initialized with a more complete context such as full environment configurations or randomized setups, allowing it to better approximate the latent dynamics of the environment.

\paragraph{Interactive Process}
Together, the agent and world model form an iterative process:
\begin{equation}
S_n', R_n' = \mathcal{W}\big( \mathcal{A}(S_0, (T_i, A_i, S_i')_{i=1}^{n-1}) \big),
\label{eq:interaction}
\end{equation}
which unrolls into a multi-turn textual trajectory generated within the world model:
\begin{equation}
\tau_{\text{wm}} = \{ S_0, T_1, A_1, S_1', \dots, T_T, A_T, S_T' \}.
\label{eq:wm_traj}
\end{equation}
Correspondingly, the real environment produces the trajectory
\begin{equation}
\tau_{\text{real}} = \{ S_0, T_1, A_1, S_1, \dots, T_T, A_T, S_T \},
\label{eq:real_traj}
\end{equation}
which serves as the reference for evaluating the fidelity and consistency of $\mathcal{W}$.

By formulating text-based environments as multi-turn interactive processes, the world model can be prompted with few-shot examplars or trained on real trajectories $\tau_{\text{real}}$ to predict next-state transitions. This formulation enables $\mathcal{W}$ to capture long-horizon dependencies and cumulative effects across interaction steps.
While prior works \citep{wang2024language,xie2024making,yang-etal-2025-efficient} primarily focus on next-state prediction accuracy, we explicitly model, train and evaluate the world model’s long-horizon consistency, which is critical for applications such as data synthesis, test-time simulator, and model-based reinforcement learning.

\subsection{Text-based Environments}

To examine the range of knowledge and dynamics required for text-based world modeling in a broad way, we adopt five representative environments spanning both structured and open-ended settings. The structured environments ALFWorld \citep{shridhar2021alfworldaligningtextembodied}, SciWorld \citep{wang2022scienceworldagentsmarter5th}, and TextWorld \citep{cote18textworld} feature bounded state spaces.
They provide deterministic or rule-governed transitions grounded in embodied, scientific, or narrative regularities.
In contrast, the open-ended environments WebShop \citep{yao2023webshopscalablerealworldweb} and StableToolBench \citep{guo2025stabletoolbenchstablelargescalebenchmarking} exhibit broad, compositional, and open-world dynamics, with diverse entities and flexible task formulations that require stronger generalization beyond fixed schemas. Table~\ref{tab:envs} in Appendix~\ref{sec:appendix_implement} summarizes these environments and their key characteristics, with examples in Appendix~\ref{sec:appendix_task_examples_and_case_studies}. Together, these settings provide a comprehensive and diversified testbed for evaluating language models as text-based world simulators.

%% file: contents/4_evaluation.tex
\section{World Model Training and Evaluation}

We summarize the world model training and evaluation setup and defer full implementation details to Appendix~\ref{sec:appendix_implement}.
Unless otherwise specified, all experiments follow the default settings described in this section.

\paragraph{Data}
We collect interaction trajectories using GPT-4o as the behavior policy. To match environment complexity (see Section~\ref{sec:data_scaling}), we gather 40K trajectories each for ALFWorld, SciWorld, and TextWorld, and 70K for WebShop. We retain both successful and failed episodes to broaden behavioral coverage for world model training.\footnote{The success/failure mixture is induced by GPT-4o’s native success rate without additional filtering.} For StableToolBench, we use the public single-turn dataset with 160K samples.

\paragraph{Initialization Context}
For ALFWorld and SciWorld, the world model is provided with full initial state descriptions (see Figures~\ref{fig:init_state_alfworld} and \ref{fig:init_state_sciworld} in Appendix~\ref{sec:appendix_world_model_initialization_context}).
In contrast, TextWorld does not expose complete initial states, and WebShop/StableToolBench are inherently partially observable. This setting places greater demands on history-based state tracking and prior knowledge to infer unobserved state variables.

\paragraph{Finetuning Models}
We use Qwen2.5-7B (base) and Llama-3.1-8B (base) as backbone models for text-based world modeling. A comparison across different model sizes is provided in Section~\ref{sec:model_scaling}.

\paragraph{Training}
Each trajectory is formatted as a multi-turn dialogue of alternating agent actions and environment responses (see Eq.~\ref{eq:real_traj}). During supervised fine-tuning, the world model predicts the next environment response conditioned on the dialogue history and the current action.

\paragraph{Metrics}

We evaluate world models along two dimensions: one-step prediction fidelity and multi-step rollout consistency.
\textbf{Fidelity.} We compute exact-match (EM) accuracy by conditioning on a real trajectory prefix $\{S_0, (A_i, S_i)_{i=1}^{n-1}, A_n\}$ and predicting the next state and reward $(S_n', R_n')$. A prediction is correct if $(S_n', R_n')$ matches the ground-truth $(S_n, R_n)$. For TextWorld, EM is a conservative lower bound since multiple surface forms can describe the same underlying state. For StableToolBench, whose outputs are highly open-ended, we additionally report word-level F1.
\textbf{Consistency.} We report: (1) \textbf{Real}: success rate in the real environment; (2) \textbf{WM}: success rate inside the world model; (3) \textbf{W2R}: success rate when replaying WM actions in the real environment; and (4) \textbf{Consistency Ratio}: $\text{CR}=\text{W2R}/\text{Real}$, where higher values indicate better long-horizon transfer (CR may exceed 1 when world model rollouts are more successful than real world).

%% file: contents/5_fidelity.tex
\section{Fidelity \& Consistency}

\subsection{Next-state Prediction Fidelity}

Table~\ref{tab:one_step_fidelity} demonstrates that pretrained LLMs exhibit meaningful in-context world modeling ability.
Models such as Gemini-2.5-flash and Claude-sonnet-4.5 achieve strong next-state prediction in structured environments like ALFWorld and SciWorld, where a handful of demonstrations provides substantial improvements (e.g., Claude rises from 56.83 to 73.08 accuracy on SciWorld with only three examples).
This suggests that contemporary LLMs encode latent knowledge of environment dynamics and can rapidly adapt their transition rules with minimal supervision.
However, these capabilities do not fully transfer to open-ended settings such as WebShop, where few-shot prompting plateaus around mid-50s, indicating that implicit world knowledge alone is insufficient for generating unconstrained, context-dependent state updates.

Supervised fine-tuning yields substantial improvements.
Open-source models trained directly on transition trajectories achieve 99\%/98\% accuracy on ALFWorld and SciWorld and reach 49\% F1 on StableToolBench.
These results indicate that robust world modeling requires dynamics-aligned training: prompting alone cannot capture the full diversity of transition patterns, whereas supervised fine-tuning enables even relatively small models to internalize them effectively.

\begin{compactgraybox}
    \textbf{Findings 1: [Short-term Fidelity]} LLMs exhibit internal latent dynamics that support in-context world modeling, while high-fidelity transition modeling depends on dynamics-aligned fine-tuning.
\end{compactgraybox}

\input{tables/next_state_accuracy.tex}
\input{tables/rollout_consistency.tex}

\subsection{Rollout Consistency}

A reliable world model requires not only high single-step prediction accuracy, but more critically, the ability to maintain consistency during extended interactions with agents. We examine two key dimensions: (1) whether small local errors compound into significant failures over long-horizon rollouts, and (2) whether the world model generalizes across different agent behaviors beyond its training distribution. Table~\ref{tab:rollout_consistency} reports consistency metrics across four environments and multiple agents; StableToolBench is omitted due to its single-turn nature.

\paragraph{Consistency Across Environments}
World models largely preserve single-step fidelity in long-horizon rollouts, especially in structured environments. In ALFWorld, SciWorld, and TextWorld, the fine-tuned Qwen2.5 world model attains high consistency ratios of 96\%, 91\%, and 92\%, indicating that multi-step trajectories generated within the world model remain executable when transferred to the real environment.
WebShop, however, exhibits lower consistency (typically below 80\%), primarily due to its open-ended nature and diverse search results that the world model struggles to simulate accurately.
This error can be substantially mitigated by grounding model rollouts with real observations. When the rollout is initialized with real search results, the consistency with GPT-4o agent increases dramatically from 56\% to nearly 100\%, demonstrating that partial real-environment anchoring effectively reduces simulation drift.

\begin{compactgraybox}
    \textbf{Findings 2: [Long-term Consistency]} World models maintain consistent long-horizon rollouts in well-structured domains, but tend to drift in open-ended environments due to high diversity, necessitating anchoring to real-world signals.
\end{compactgraybox}

\paragraph{How does behavior shift affect consistency?}
Beyond environment-specific factors, world model consistency also depends on how well agent behaviors match the training distribution. Lower-capacity agents such as GPT-4o-mini yield consistency ratios frequently below 70\%, whereas stronger agents like GPT-4.1, GPT-5, and Claude reliably exceed 90\%. This disparity stems from weaker agents taking actions misaligned with task objectives, causing their trajectories to drift outside the training distribution. In contrast, higher-capacity agents preserve goal-directed behavior that aligns with the expert policy (GPT-4o) used for trajectories sampling, enabling higher consistency.
These results highlight the importance of diversifying training trajectories rather than relying solely on a single strong agent, as further discussed in Section \ref{sec:agent_mix_training}.

%% file: tables/next_state_accuracy.tex
\begin{table}[]
    \centering
    \resizebox{0.6\columnwidth}{!}{
    \begin{tabular}
        {
            l@{}
            S[table-format=3.3]@{} S[table-format=3.3]@{}
            S[table-format=3.3]@{} S[table-format=3.3]@{}
            S[table-format=3.3]@{} S[table-format=3.3]
        }
    \toprule
    \textbf{Environment} & \textbf{AW} & \textbf{SW} & \textbf{TW} & \textbf{WS} & \textbf{STB} & \textbf{STB$_{\text{F1}}$} \\
    \midrule\addlinespace[-0.000ex]
    \rowcolor{gray!12}
    \multicolumn{7}{c}{\textbf{\textit{Zero-shot}}} \\[-0.4ex]
    \midrule
    GPT-4o-mini          & 45.20       & 40.68       & 0.36        & 56.59       & 0.00 & 13.94         \\
    GPT-4o               & 44.45       & 45.78       & 7.86        & 58.20       & 0.00 & 11.88         \\
    GPT-4-turbo          & 42.64       & 34.14       & 0.00        & 52.45       & 0.00 & 12.64         \\
    GPT-4.1              & 43.56       & 35.65       & 0.00        & 58.07       & 0.00 & 12.83         \\
    GPT-5                & 35.09       & 13.06       & 9.20        & 46.12       & 0.00 & 8.02         \\
    Gemini-2.5-flash     & 50.00       & 44.81       & 3.51       & 57.64        & 0.00 & 8.74         \\
    Claude-sonnet-4.5    & 64.73       & 56.83       & 17.70       & 58.80       & 0.00 & 11.36         \\
    \midrule\addlinespace[-0.000ex]
    \rowcolor{gray!12}
    \multicolumn{7}{c}{\textbf{\textit{Few-shot (3 shot)}}} \\[-0.4ex]
    \midrule
    GPT-4o-mini         & 63.79       & 56.26       & 11.43       & 61.93       & 0.00 & 13.44         \\
    GPT-4o              & 56.88       & 48.98       & 14.11       & 64.62       & 0.00 & 11.08         \\
    GPT-4-turbo         & 62.56       & 50.08        & 11.66      & 62.76       & 0.00 & 10.72         \\
    GPT-4.1             & 63.37       & 51.56       & 13.39       & 64.23       & 0.00 & 10.33         \\
    GPT-5               & 67.13       & 49.44       & 44.27       & 65.90       & 0.00 & 6.28         \\
    Gemini-2.5-flash    & 61.85       & 61.20       & 40.35       & 66.09       & 0.00 & 8.47         \\
    Claude-sonnet-4.5   & 77.04       & 73.08       & 49.12       & 56.65       & 0.00 & 13.11         \\
    \midrule\addlinespace[-0.000ex]
    \rowcolor{gray!12}
    \multicolumn{7}{c}{\textbf{\textit{SFT}}} \\[-0.4ex]
    \midrule
    Qwen2.5-7B           & 99.87       & 98.60       & 70.60       & 79.05       & 48.90 & 79.15        \\
    Llama3.1-8B          & 99.71       & 98.64       & 70.45       & 77.24       &  49.25 & 78.97   \\
    \bottomrule
    \end{tabular}
    }
    \caption{Next-state prediction EM accuracy (\%) of prompt-based and finetuned models across five environments. AW, SW, TW, WS and STB denote ALFWorld, SciWorld, TextWorld, WebShop and StableToolBench, respectively. STB$_{\text{F1}}$ denotes the word-level F1 score for StableToolBench, given its open-ended output space.}
    \label{tab:one_step_fidelity}
    \vspace{-15pt}
\end{table}

%% file: tables/rollout_consistency.tex
\begin{table*}[t]
    \centering
    \resizebox{\textwidth}{!}{
        \begin{tabular}{
        l@{}
        S[table-format=3.3]@{} S[table-format=3.3]@{} S[table-format=3.3]@{} S[table-format=2.3]@{}
        S[table-format=3.3]@{} S[table-format=3.3]@{} S[table-format=3.3]@{} S[table-format=2.3]@{}
        S[table-format=4.3]@{} S[table-format=3.3]@{} S[table-format=3.3]@{} S[table-format=2.3]@{}
        S[table-format=3.3]@{} S[table-format=3.3]@{} S[table-format=3.3]@{} S[table-format=2.3]@{}
        }
        \toprule\addlinespace[-0.000ex]
        \multirow{2}{*}{\textbf{Agent}} &
        \multicolumn{4}{c}{\cellcolor{red!6}ALFWorld} &
        \multicolumn{4}{c}{\cellcolor{LavenderLight!20}SciWorld} &
        \multicolumn{4}{c}{\cellcolor{orange!10}TextWorld} &
        \multicolumn{4}{c}{\cellcolor{violet!8}WebShop} \\[-0.4ex]
        & \multicolumn{1}{c}{\cellcolor{red!6}Real} & \multicolumn{1}{c}{\cellcolor{red!6}WM} & \multicolumn{1}{c}{\cellcolor{red!6}W2R} & \multicolumn{1}{c}{\cellcolor{red!6}CR}
        & \multicolumn{1}{c}{\cellcolor{LavenderLight!20}Real} & \multicolumn{1}{c}{\cellcolor{LavenderLight!20}WM} & \multicolumn{1}{c}{\cellcolor{LavenderLight!20}W2R} & \multicolumn{1}{c}{\cellcolor{LavenderLight!20}CR}
        & \multicolumn{1}{c}{\cellcolor{orange!10}Real} & \multicolumn{1}{c}{\cellcolor{orange!10}WM} & \multicolumn{1}{c}{\cellcolor{orange!10}W2R} & \multicolumn{1}{c}{\cellcolor{orange!10}CR}
        & \multicolumn{1}{c}{\cellcolor{violet!8}Real} & \multicolumn{1}{c}{\cellcolor{violet!8}WM} & \multicolumn{1}{c}{\cellcolor{violet!8}W2R} & \multicolumn{1}{c}{\cellcolor{violet!8}CR} \\[-0.6ex]
        \midrule[-0.4ex]\midrule\addlinespace[-0.000ex]
        \rowcolor{gray!12}
        \multicolumn{17}{c}{
            \rule{0pt}{1.1em}
            \textbf{\textit{Qwen2.5-7B WorldModel}}
            \rule[-0.3em]{0pt}{1.1em}
        } \\
        [-0.4ex]\midrule\addlinespace[-0.000ex]

        GPT-4o-mini  & 7.69 & 7.69 & 7.69 & \colorbox{MintGreen!100}{1.00}
                    & 12.64 & 12.04 & 8.90 & \colorbox{MintGreen!50}{0.70}
                    & 97.44 & 100.00 & 69.36 & \colorbox{MintGreen!52}{0.71}
                    & 5.99 & 4.85 & 0.97 & \colorbox{MintGreen!0}{0.16} \\

        GPT-4o       & 58.00 & 55.90 & 57.44 & \colorbox{MintGreen!98}{0.99}
                    & 34.97 & 37.63 & 31.44 & \colorbox{MintGreen!83}{0.90}
                    & 98.84 & 100.00 & 96.53 & \colorbox{MintGreen!97}{0.98}
                    & 29.36 & 17.43 & 16.51 & \colorbox{MintGreen!27}{0.56} \\

        GPT-4-turbo  & 74.21 & 62.56 & 64.62 & \colorbox{MintGreen!78}{0.87}
                    & 36.79 & 50.00 & 36.60 & \colorbox{MintGreen!98}{0.99}
                    & 100.00 & 99.42 & 98.84 & \colorbox{MintGreen!98}{0.99}
                    & 17.73 & 14.89 & 11.70 & \colorbox{MintGreen!43}{0.66} \\

        GPT-4.1      & 67.20 & 68.56 & 69.59 & \colorbox{MintGreen!100}{1.04}
                    & 43.41 & 45.79 & 46.32 & \colorbox{MintGreen!100}{1.07}
                    & 100.00 & 100.00 & 100.00 & \colorbox{MintGreen!100}{1.00}
                    & 21.14 & 12.22 & 12.22 & \colorbox{MintGreen!30}{0.58} \\

        GPT-5        & 91.00 & 84.62 & 86.67 & \colorbox{MintGreen!92}{0.95}
                    & 68.21 & 64.10 & 61.03 & \colorbox{MintGreen!82}{0.89}
                    & 100.00 & 100.00 & 100.00 & \colorbox{MintGreen!100}{1.00}
                    & 51.00 & 33.03 & 31.19 & \colorbox{MintGreen!35}{0.61} \\

        Gemini-2.5-flash & 50.50 & 51.79 & 52.31 & \colorbox{MintGreen!100}{1.04}
                        & 56.00 & 39.49 & 45.64 & \colorbox{MintGreen!70}{0.82}
                        & 100.00 & 100.00 & 76.30 & \colorbox{MintGreen!60}{0.76}
                        & 25.00 & 21.10 & 18.35 & \colorbox{MintGreen!55}{0.73} \\

        Claude-sonnet-4.5 & 82.00 & 76.00 & 76.00 & \colorbox{MintGreen!88}{0.93}
                        & 66.00 & 45.64 & 57.95 & \colorbox{MintGreen!80}{0.88}
                        & 100.00 & 100.00 & 100.00 & \colorbox{MintGreen!100}{1.00}
                        & 61.00 & 49.00 & 50.00 & \colorbox{MintGreen!70}{0.82} \\

        \textbf{Average}  & 61.51 & 58.16 & 59.19 & \colorbox{MintGreen!93}{0.96}
                        & 45.43 & 42.10 & 41.13 & \colorbox{MintGreen!85}{0.91}
                        & 99.47 & 99.92 & 91.58 & \colorbox{MintGreen!87}{0.92}
                        & 30.17 & 21.79 & 20.13 & \colorbox{MintGreen!45}{0.67} \\

        \addlinespace[-0.000ex]\midrule\addlinespace[-0.000ex]
        \rowcolor{gray!12}
        \multicolumn{17}{c}{
            \rule{0pt}{1.1em}
            \textbf{\textit{Llama3.1-8B WorldModel}}
            \rule[-0.3em]{0pt}{1.1em}
        } \\
        [-0.4ex]\midrule\addlinespace[-0.000ex]
        GPT-4o-mini  & 7.69 & 9.74 & 9.74 & \colorbox{MintGreen!100}{1.27}
                    & 12.64 & 10.78 & 8.21 & \colorbox{MintGreen!42}{0.65}
                    & 97.44 & 92.48 & 57.80 & \colorbox{MintGreen!32}{0.59}
                    & 5.99 & 2.75 & 0.00 &  \colorbox{MintGreen!0}{0.00} \\

        GPT-4o       & 58.00 & 58.46 & 56.92 & \colorbox{MintGreen!97}{0.98}
                    & 34.97 & 37.63 & 32.99 & \colorbox{MintGreen!90}{0.94}
                    & 98.84 & 97.11 & 90.17 & \colorbox{MintGreen!85}{0.91}
                    & 29.36 & 23.81 & 22.62 & \colorbox{MintGreen!62}{0.77} \\

        GPT-4-turbo  & 74.21 & 67.53 & 67.01 & \colorbox{MintGreen!83}{0.90}
                    & 36.79 & 52.31 & 44.10 & \colorbox{MintGreen!100}{1.20}
                    & 100.00 & 97.69 & 93.06 & \colorbox{MintGreen!88}{0.93}
                    & 17.73 & 25.47 & 17.92 & \colorbox{MintGreen!100}{1.01} \\

        GPT-4.1      & 67.20 & 68.72 & 68.21 &  \colorbox{MintGreen!100}{1.02}
                    & 43.41 & 45.13 & 35.38 &  \colorbox{MintGreen!70}{0.82}
                    & 100.00 & 98.27 & 94.22 &  \colorbox{MintGreen!90}{0.94}
                    & 21.14 & 19.27 & 17.43 & \colorbox{MintGreen!70}{0.82} \\

        GPT-5        & 91.00 & 82.56 & 81.54 & \colorbox{MintGreen!83}{0.90}
                    & 68.21 & 63.07 & 57.44 & \colorbox{MintGreen!73}{0.84}
                    & 100.00 & 98.84 & 94.80 & \colorbox{MintGreen!92}{0.95}
                    & 51.00 & 31.19 & 30.28 & \colorbox{MintGreen!32}{0.59} \\

        Gemini-2.5-flash & 50.50 & 53.33 & 53.33 & \colorbox{MintGreen!100}{1.06}
                        & 56.00 & 57.95 & 52.31 & \colorbox{MintGreen!88}{0.93}
                        & 100.00 & 99.42 & 93.06 & \colorbox{MintGreen!88}{0.93}
                        & 25.00 & 22.02 & 17.43 &  \colorbox{MintGreen!50}{0.70} \\

        Claude-sonnet-4.5 & 82.00 & 84.00 & 84.00 & \colorbox{MintGreen!100}{1.02}
                        & 66.00 & 58.46 & 53.33 & \colorbox{MintGreen!68}{0.81}
                        & 100.00 & 93.33 & 90.00 & \colorbox{MintGreen!83}{0.90}
                        & 61.00 & 60.00 & 55.00 & \colorbox{MintGreen!83}{0.90} \\

        \textbf{Average}  & 61.51 & 60.62 & 60.11 & \colorbox{MintGreen!97}{0.98}
                        & 45.43 & 46.48 & 30.52 & \colorbox{MintGreen!82}{0.89}
                        & 99.47 & 96.73 & 87.59 & \colorbox{MintGreen!80}{0.88}
                        & 30.17 & 26.36 & 22.95 & \colorbox{MintGreen!60}{0.76} \\

        \addlinespace[-0.000ex]\bottomrule
        \end{tabular}
    }
\caption{Task success rate (\%) of different agents across four environments. ``Real'', ``WM'', and ``W2R'' denote the success rate under real environment, world model, and world model-to-real execution. The last column reports the consistency ratio (CR=W2R/Real), with higher values (darker green color) indicating better rollout fidelity.}
\label{tab:rollout_consistency}
\end{table*}

%% file: contents/6_scalability.tex
\section{Scalability \& Robustness}\label{sec:scaling_generalization}

\subsection{Data Scaling Laws for World Models}\label{sec:data_scaling}

\begin{figure*}
    \centering
    \includegraphics[width=1\textwidth]{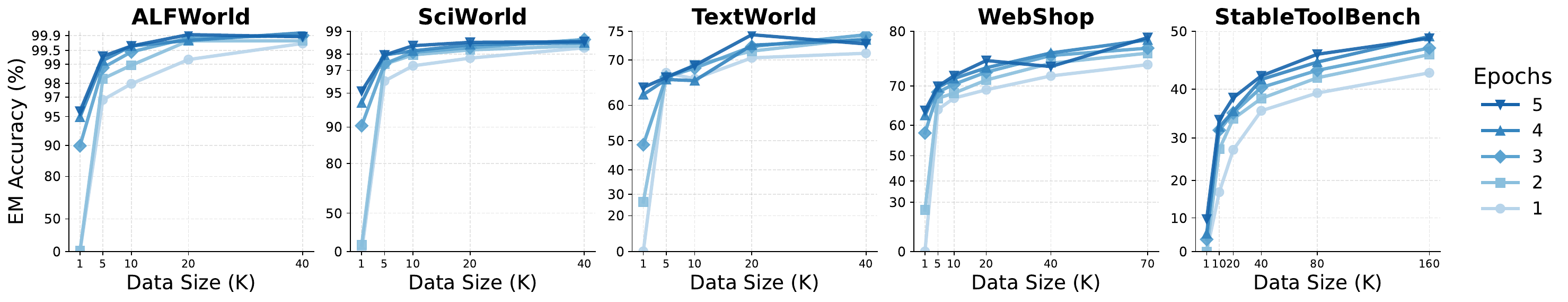}
    \caption{Next-state prediction accuracy under varying training data sizes on Qwen2.5-7B. Structured settings saturate with modest data (\textasciitilde 20K), whereas open-ended settings continue to benefit from larger datasets.
    \newline\footnotesize
    \textbf{Note.} We apply a nonlinear y-axis transform
    \(f(y) = 100 - 20 \log_{10}(\max(100 - y, 0.01) + 1)\)
    to better reveal growth trends.}
    \label{fig:data_size}
\end{figure*}

\begin{figure*}
    \centering
    \includegraphics[width=1\textwidth]{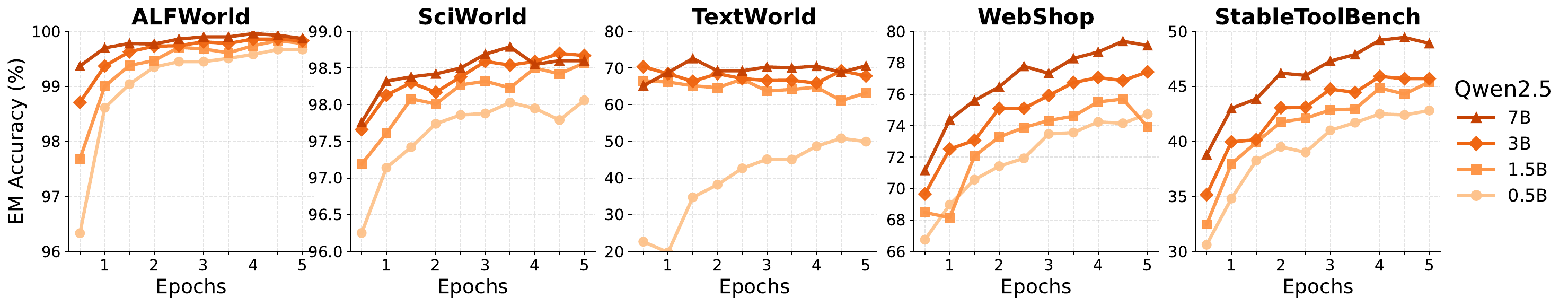}
    \caption{Next-state prediction accuracy on Qwen2.5 family. Smaller models (\textasciitilde1.5B) capture structured dynamics effectively, whereas more complex settings benefit markedly from increased model capacity.}
    \label{fig:model_size}
\end{figure*}

To investigate how world model performance scales with data, we vary training trajectories from 1K to 160K and evaluate single-step accuracy. As shown in Figure~\ref{fig:data_size}, structured environments (ALFWorld, SciWorld, TextWorld) improve rapidly and saturate around 20K trajectories, consistent with their low-entropy, rule-driven dynamics. In contrast, open-ended environments scale more gradually: WebShop benefits from additional data up to roughly 70K trajectories, while StableToolBench shows no saturation at 160K samples due to long-tail linguistic variation and highly compositional API behaviors. These results indicate that world modeling exhibits environment-dependent scaling: structured environments are highly data-efficient, whereas open-ended domains require substantially larger datasets.

\subsection{Model Size Effects}\label{sec:model_scaling}

We next analyze how model capacity shapes world model performance (Figure~\ref{fig:model_size}). Mirroring data-scaling trends, model size interacts strongly with environment complexity. In structured environments, performance saturates quickly: 1.5B models already capture core transition dynamics, with further scaling yielding only marginal improvements. In open-ended environments, however, capacity matters substantially. Smaller models struggle to represent rich linguistic variability and compositional tool usage, whereas larger models offer steady accuracy gains. Together with the data-scaling results, these findings indicate that success in open-ended world modeling requires both extensive trajectories and sufficient model capacity to internalize long-tailed, high-entropy dynamics.

\begin{compactgraybox}
\textbf{Findings 3: [LLMs are Scalable World Models]} World modeling performance scales systematically with data volume, model size and environment complexity.
\end{compactgraybox}

\subsection{Beyond Memorized Environments}

\begin{figure}[t]
    \centering
    \includegraphics[width=0.5\columnwidth]{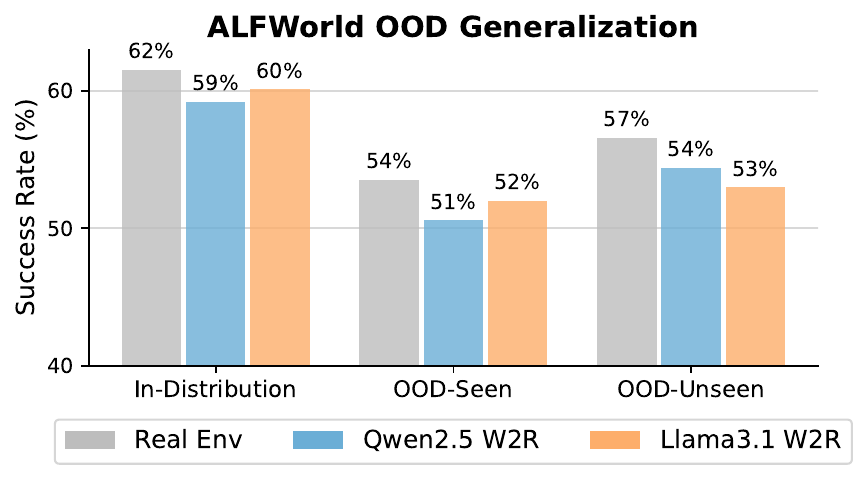}
    \caption{Task success rate (\%) in ALFWorld under different OOD settings. Success rate averaged over different agents, with full results provided in Table~\ref{appendix:tab:ood_generalization} of Appendix~\ref{sec:appendix_full_results}. World models maintain strong performance even when layouts or room types change.}
    \label{fig:ood_generalization}
\end{figure}

A central question in world model design is how well they generalize across unseen settings. Using ALFWorld as a representative case, we analyze two out-of-distribution test splits following the original environment settings \citep{shridhar2021alfworldaligningtextembodied}: \textit{OOD-Seen}, which keeps the room type but alters the layout, and \textit{OOD-Unseen}, which introduces entirely new room types or unseen layout configurations.
As shown in Figure~\ref{fig:ood_generalization}, the world models maintain success rates closely aligned with the real environment across both OOD settings even when the spatial configuration shifts or novel room types appear. These results indicate that the LLM world model captures transferable transition dynamics rather than memorizing specific layouts, demonstrating strong robustness to structural variations in environment state space.

\subsection{Cross-Env Transfer via Joint Training}\label{sec:mix_environment_training}

\begin{figure*}[!htb]
    \centering
    \includegraphics[width=1\textwidth]{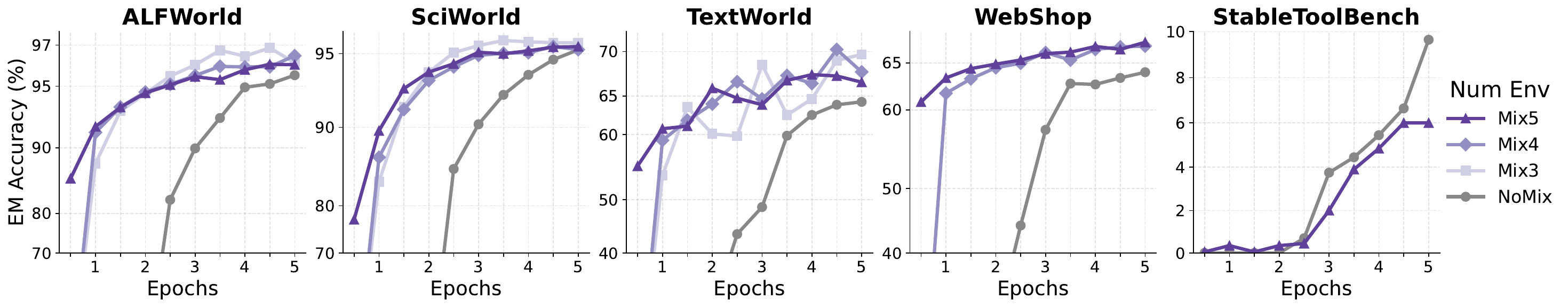}
    \caption{Next-state prediction accuracy under mixed and separate training on Qwen2.5-7B, with 1K samples per environment. We begin by mixing structured environments (ALFWorld, SciWorld, TextWorld) and then progressively incorporate open-ended environments (WebShop, StableToolBench), yielding the Mix3, Mix4, and Mix5 settings.}
    \label{fig:mix_training}
\end{figure*}

Training world models in isolation often limits their ability to generalize beyond a single environment, motivating us to investigate whether jointly training on multiple environments can yield transferable gains. We therefore evaluate three mixed-training configurations: Mix3 (ALFWorld, SciWorld, TextWorld), Mix4 (with WebShop), and Mix5 (with StableToolBench), allocating 1K trajectories per environment to match the data budget of individually trained models. As shown in Figure~\ref{fig:mix_training}, mixed training consistently accelerates learning and improves final accuracy, with particularly strong gains in TextWorld and WebShop, suggesting that the model effectively internalizes and reuses shared physical, procedural, and narrative dynamics across tasks. The exception is StableToolBench, whose schema-centric, single-turn structure is underrepresented in the mixture, causing separately trained model to outperform. Overall, these results show that mixed data provides stable positive gains and, importantly, enables practical deployments where a single world model can robustly serve multiple environments.

\subsection{Behavioral Coverage for Robust World Modeling}\label{sec:agent_mix_training}

\input{tables/mix_agent_training.tex}

As behavior shifts reduce consistency, we ask whether broader behavioral coverage improves generalization. We train a world model on mixed-agent trajectories and compare it to a GPT-4o-only baseline. Table~\ref{tab:mix_agent_training} shows marked OOD gains for weaker agents: GPT-4o-mini’s consistency ratio rises from 0.49 to 0.81, and GPT-4-turbo also improves. This indicates that expert-only trajectories are insufficient under distribution shift; incorporating diverse agent behaviors is crucial for improving generalization and rollout stability.

\begin{compactgraybox}
\textbf{Findings 4: [World Models Learn Generalizable Dynamics]} World models generalize beyond memorized configurations and transfer across environments and agent policies, where coverage of environments and diversity of agent behaviors play critical roles.
\end{compactgraybox}

%% file: tables/mix_agent_training.tex
\begin{table}[t]
    \centering
    \resizebox{0.6\columnwidth}{!}{
    \begin{tabular}{
    l
    c
    S[table-format=2.2] S[table-format=2.2] S[table-format=1.2]
    S[table-format=2.2] S[table-format=2.2] S[table-format=1.2]
    }
    \toprule\addlinespace[-0.000ex]
    \multirow{2}{*}{\textbf{Agent}} &
    \multicolumn{1}{c}{\cellcolor{gray!12}} &
    \multicolumn{3}{c}{\cellcolor{red!6}Single Agent Traj} &
    \multicolumn{3}{c}{\cellcolor{LavenderLight!20}Mix Agent Traj} \\
    & \multicolumn{1}{c}{\cellcolor{gray!12}\multirow{-2}{*}{Real}} &
    \multicolumn{1}{c}{\cellcolor{red!6}WM} &
    \multicolumn{1}{c}{\cellcolor{red!6}W2R} &
    \multicolumn{1}{c}{\cellcolor{red!6}CR} &
    \multicolumn{1}{c}{\cellcolor{LavenderLight!20}WM} &
    \multicolumn{1}{c}{\cellcolor{LavenderLight!20}W2R} &
    \multicolumn{1}{c}{\cellcolor{LavenderLight!20}CR} \\
    \addlinespace[-0.000ex]\midrule\addlinespace[-0.000ex]

    Qwen3-235B & 24.00 & 24.00 & 18.00 & \cellcolor{MintGreen!75}{0.75} & 26.00 & 18.00 & \cellcolor{MintGreen!75}{0.75} \\
    GPT-4o & 34.97 & 32.31 & 26.67 & \cellcolor{MintGreen!60}{0.76} & 32.31 & 26.67 & \cellcolor{MintGreen!60}{0.76} \\
    GPT-5 & 68.21 & 55.38 & 59.49 & \cellcolor{MintGreen!75}{0.87} & 69.74 & 60.00 & \cellcolor{MintGreen!90}{0.88} \\
    Claude-sonnet-4.5 & 66.00 & 40.51 & 57.44 & \cellcolor{MintGreen!75}{0.87} & 52.31 & 49.74 & \cellcolor{MintGreen!60}{0.75} \\
    \textbf{ID Agent Average} & 48.30 & 38.05 & 40.40 & \cellcolor{MintGreen!75}{0.81} & 45.09 & 38.60 & \cellcolor{MintGreen!75}{0.79} \\
    \addlinespace[-0.000ex]\midrule\addlinespace[-0.000ex]

    GPT-4o-mini & 12.64 & 5.64 & 6.15 & \cellcolor{MintGreen!30}{0.49} & 13.39 & 10.26 & \cellcolor{MintGreen!75}{0.81} \\
    GPT-4-turbo & 36.79 & 32.31 & 38.97 & \cellcolor{MintGreen!100}{1.06} & 51.28 & 42.56 & \cellcolor{MintGreen!100}{1.16} \\
    GPT-4.1 & 43.41 & 28.72 & 36.41 & \cellcolor{MintGreen!60}{0.84} & 52.31 & 36.41 & \cellcolor{MintGreen!60}{0.84} \\
    Gemini-2.5-flash & 56.00 & 36.92 & 51.79 & \cellcolor{MintGreen!75}{0.92} & 56.92 & 45.64 & \cellcolor{MintGreen!75}{0.82} \\
    \textbf{OOD Agent Average} & 37.21 & 25.90 & 33.33 & \cellcolor{MintGreen!60}{0.83} & 43.48 & 33.72 & \cellcolor{MintGreen!75}{0.91} \\

    \addlinespace[-0.000ex]\bottomrule
    \end{tabular}
    }
\caption{Task success rate (\%) in SciWorld under different training data compositions. ``Single Agent Traj'' uses only 4K GPT-4o trajectories for training, whereas ``Mix Agent Traj'' combines trajectories from ID agents, with 1K trajectories from each.}
\label{tab:mix_agent_training}
\end{table}

%% file: contents/7_agent_utility.tex
\section{Agent Utility}\label{sec:applications}

\input{tables/double_check.tex}

In real-world decision-making, some actions are irreversible and costly, creating a safety bottleneck: a single mistaken commitment can end an episode or cause unrecoverable loss. This motivates using world models as a \emph{rewindable imagined world} to evaluate high-stakes actions before execution. WebShop exemplifies this setting: once the agent checks out, the episode ends and errors cannot be undone. We therefore use the world model as a lightweight pre-execution verifier. Before committing to checkout, the agent simulates the outcome; it executes the action only when the prediction indicates success, otherwise it continues interacting with the environment. We vary the verification budget (0, 2, 4, 10, 50).

As shown in Table~\ref{tab:double_check}, verification improves success rates for all agents, with the largest gains for medium-capacity models. However, returns are not monotonic, since repeated verification changes the trajectory context and shifts the agent’s action distribution, inducing distribution shift that can weaken alignment between imagined and real outcomes. In practice, moderate budgets (e.g., 2--10 checks) provide the best trade-off, reducing irreversible failures without destabilizing behavior.

\begin{compactgraybox}
\textbf{Findings 5: [As a Rewindable Safety Verifier]} By simulating irreversible actions and gating execution on predicted success, world models help agents avoid costly commitments and achieve higher task success.
\end{compactgraybox}

\subsection{Synthetic Data Competes with Real}

\begin{figure}[t]
    \centering
    \begin{minipage}[t]{0.49\columnwidth}
        \centering
        \includegraphics[width=\linewidth]{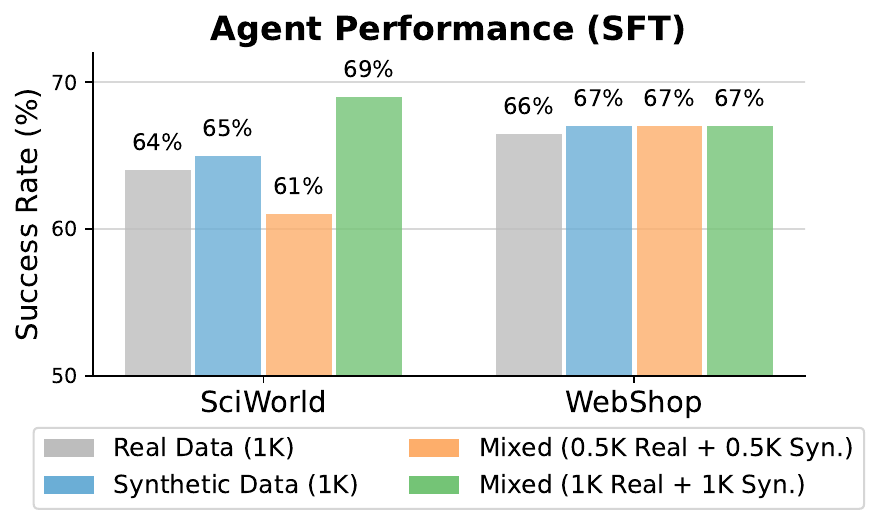}
        \captionof{figure}{Task success rate (\%) of Qwen2.5-7B-Instruct SFT trained agents with different data synthesis strategies in SciWorld and WebShop.}
        \label{fig:data_synthesis}
    \end{minipage}\hfill
    \begin{minipage}[t]{0.49\columnwidth}
        \centering
        \includegraphics[width=\linewidth]{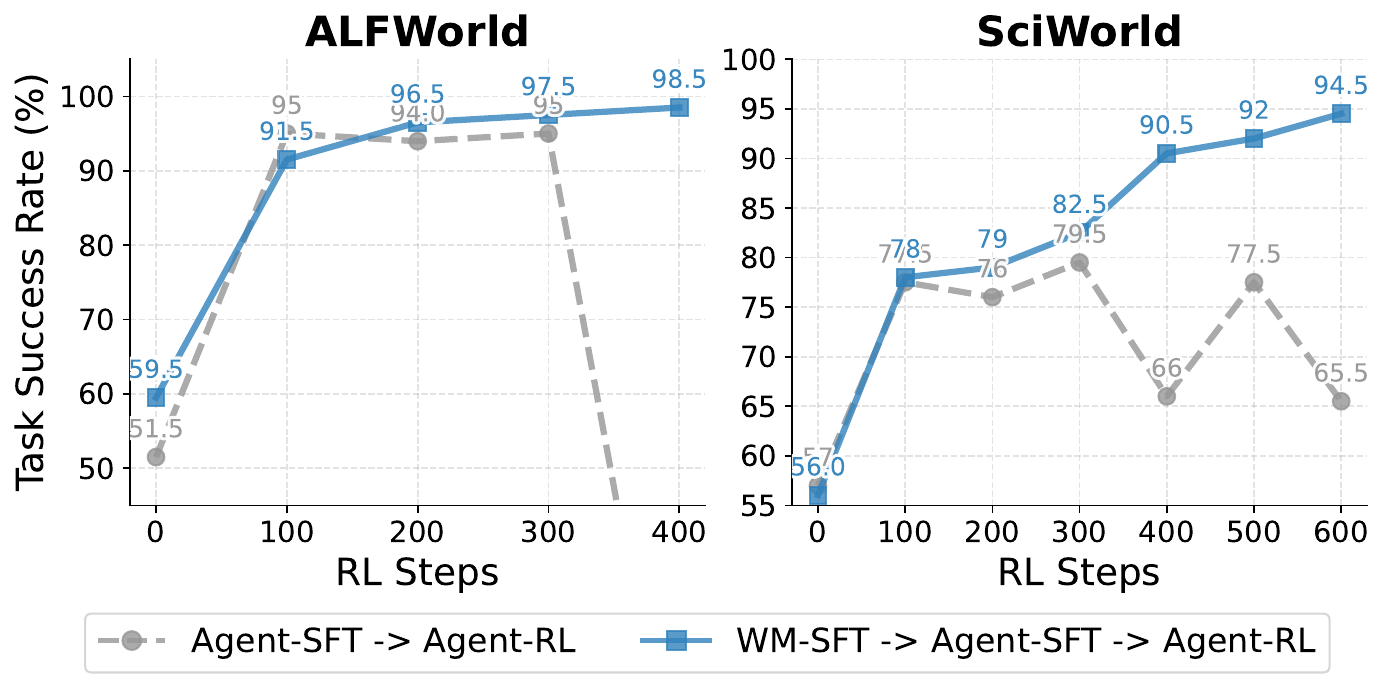}
        \captionof{figure}{Task success rate (\%) of Qwen2.5-7B-Instruct RL trained agents with and without early experience in ALFWorld and SciWorld.}
        \label{fig:early_experience}
    \end{minipage}
\end{figure}

When real interaction is expensive, slow, or constrained, agents face an experience bottleneck. A world model can potentially alleviate this bottleneck by synthesizing trajectories that substitute for a portion of real experience. To examine this, we collect 1{,}000 successful trajectories from either the real environment or the world model\footnote{{Implementation details in Appendix~\ref{sec:appendix_syn_vs_real}}}, and construct four SFT datasets for Qwen2.5-7B-Instruct: Real 1K, Syn 1K, 0.5K+0.5K, and a 1K+1K mixture, all trained under identical procedures.
Figure~\ref{fig:data_synthesis} shows that world model--generated trajectories are highly competitive with real data. In SciWorld, Syn 1K matches Real 1K, while the 1K+1K mixture outperforms either source alone. In WebShop, synthetic data remains similarly effective, and mixed regimes yield the most stable gains. Overall, these results suggest that synthetic experience can reduce reliance on real-environment interaction, providing an alternative pathway for scaling agent learning when real experience is limited.

\subsection{Early Experience for Policy Learning}

Recent work~\citep{zhang2025agentlearningearlyexperience} suggests that exposing a model to environment dynamics before explicit policy learning can provide a useful inductive bias: anticipating consequences may reduce unguided exploration and stabilize early RL. To study this in our text-based decision environments, we compare (1) standard Agent-SFT $\rightarrow$ RL baseline; and (2) world-model warmup pipeline (WM-SFT $\rightarrow$ Agent-SFT $\rightarrow$ RL), where the agent is first exposed to environment dynamics with same objective as world model training\footnote{Implementation details in Appendix~\ref{sec:appendix_early_experience}}.

Figure~\ref{fig:early_experience} indicates that early experience delivers consistent gains on both ALFWorld and SciWorld. By exposing the agent to environment dynamics before policy learning, early experience stabilizes RL training, reducing failures driven by incorrect commonsense priors, and ultimately yields higher final success rates than the baseline. Overall, early experience provides a promising direction for improving learning effectiveness.

\begin{compactgraybox}
    \textbf{Findings 6: [Learning from World-Model Dynamics]} By providing both scalable supervision and early dynamics exposure, world models reduce reliance on real interaction and improve learning efficiency, enabling agents to learn more effectively from experience.
\end{compactgraybox}

%% file: tables/double_check.tex
\subsection{Can World Models Prevent Irreversible Mistakes?}
\begin{table}
    \centering
    \resizebox{0.8\columnwidth}{!}{
        \begin{tabular}{lccccc}
        \toprule
        \textbf{Agent}       & 0 & 2  & 4  & 10 & 50 \\ \midrule
        GPT-4o-mini & 5.99            & 7.50 \textcolor{green!70!black}{(+1.51)}  & 7.55 \textcolor{green!70!black}{(+1.56)}  & 7.59 \textcolor{green!70!black}{(+1.60)} & 7.59 \textcolor{green!70!black}{(+1.60)}  \\
        GPT-4o      & 29.36           & 32.41 \textcolor{green!70!black}{(+3.05)} & 33.94 \textcolor{green!70!black}{(+4.58)} & 34.86 \textcolor{green!70!black}{(+5.50)} & 36.70 \textcolor{green!70!black}{(+7.34)} \\
        GPT-4-turbo & 17.73           & 33.33 \textcolor{green!70!black}{(+15.60)} & 27.05 \textcolor{green!70!black}{(+9.32)} & 29.37 \textcolor{green!70!black}{(+11.64)} & 25.60 \textcolor{green!70!black}{(+7.87)} \\
        GPT-4.1     & 21.14           & 23.59 \textcolor{green!70!black}{(+2.45)}  & 23.59 \textcolor{green!70!black}{(+2.45)}  & 23.08 \textcolor{green!70!black}{(+1.94)} & 25.13 \textcolor{green!70!black}{(+3.99)} \\
        GPT-5       & 51.00           & 53.27 \textcolor{green!70!black}{(+2.27)}  & 53.77 \textcolor{green!70!black}{(+2.77)}  & 53.27 \textcolor{green!70!black}{(+2.77)} & 51.50 \textcolor{green!70!black}{(+0.50)} \\
        Gemini-2.5-flash     & 25.00  & 31.00 \textcolor{green!70!black}{(+6.00)}   & 29.50 \textcolor{green!70!black}{(+4.50)}  & 28.00 \textcolor{green!70!black}{(+3.00)}    & 27.50  \textcolor{green!70!black}{(+1.50)} \\
        Claude-sonnet-4.5 & 61.00           & 62.00 \textcolor{green!70!black}{(+1.00)}  & 65.00 \textcolor{green!70!black}{(+4.00)}  & 64.00 \textcolor{green!70!black}{(+3.00)} & 62.00 \textcolor{green!70!black}{(+1.00)} \\
        \bottomrule
        \end{tabular}
    }
    \caption{Task success rate (\%) of different agents in WebShop with varying numbers of max pre-execution verification attempts using the world model. The numbers in parentheses indicate the improvement over the baseline without verification.}
    \label{tab:double_check}
\end{table}

%% file: contents/8_conclusion.tex
\section{Conclusion}

This work set out to investigate a simple yet far-reaching question: can the paradigm that enables large language models to model words also enable them to model worlds, and in turn support more effective agent learning from experience? Using text-based environments as a controlled testbed, we recast world modeling as multi-turn next-state prediction under interaction, and introduce a systematic framework for evaluating fidelity, scalability, and agent utility.

Our results provide strong evidence that LLMs can serve as implicit text-based world models. When trained with dynamics-aligned supervision at sufficient scale and coverage, they maintain coherent latent state over extended horizons and yield tangible benefits for downstream agents, including safer decision making, scalable experience generation, and improved learning efficiency. At the same time, these gains are not universal: robustness depends critically on behavioral coverage, distributional alignment, and environment complexity, delineating concrete regimes in which world modeling meaningfully supports agent learning.

Taken together, these findings establish an empirical foundation for treating LLMs not merely as sequence predictors, but as learned simulators of interactive worlds. By bridging next-token prediction with next-state modeling, this work points toward a unifying view of language models as world models for agents—and opens the door to extending these ideas beyond text to richer, multimodal, and embodied domains.

%% file: contents/appendix.tex
\newpage
\clearpage
\onecolumn 
\appendix

\renewcommand{\contentsname}{\Large Appendix}
{\hypersetup{linkcolor=black}
\tableofcontents
}

\addtocontents{toc}{\protect\setcounter{tocdepth}{2}}

\newpage

\section{Implementation Details}
\label{sec:appendix}

\subsection{World Model Training and Evaluation}
\label{sec:appendix_implement}

\paragraph{Environments} We evaluate five text-based environments, including ALFWorld \citep{shridhar2021alfworldaligningtextembodied}, SciWorld \citep{wang2022scienceworldagentsmarter5th}, TextWorld \citep{cote18textworld}, WebShop \citep{yao2023webshopscalablerealworldweb}, and StableToolBench \citep{guo2025stabletoolbenchstablelargescalebenchmarking}. Table~\ref{tab:envs} summarizes these environments along four dimensions: the nature of the environment, the abilities required of an agent, the form of the underlying world state, and the modeling capabilities demanded of a world model.

\input{tables/env_description.tex}

\paragraph{Data Sources and Sizes} For ALFWorld, SciWorld and WebShop, we follow the data splits provided in AgentGym\footnote{\url{https://github.com/WooooDyy/AgentGym}} \citep{xi2024agentgym}.
For TextWorld, we follow the official TextWorld repository\footnote{\url{https://github.com/microsoft/TextWorld}} to generate game files and randomly split into 2.5K training games and 200 test games.
For StableToolBench, we filtered the StableToolBench MirrorAPI dataset\footnote{\url{https://huggingface.co/datasets/stabletoolbench/MirrorAPI-Training}} and remove samples with errors or incomplete information, and use 160K API pairs for training and 2K pairs for testing. The data sizes for different environments are summarized in Table~\ref{tab:training_data_sizes}.

\paragraph{Trajectories Collection}
We utilize the AgentGym \citep{xi2024agentgym} framework to collect long-horizon interaction trajectories using GPT-4o
as the agent across four interactive environments: ALFWorld, SciWorld, TextWorld, and WebShop. We maintain consistent system prompts (Appendix~\ref{sec:appendix_system_prompts}), interaction protocols, environment configurations as in AgentGym.
The sampling temperature is set to 1.0 with Top-p of 1.0, and a maximum of 50 interaction turns per trajectory.
System prompts used for trajectory collection are provided in Figure~\ref{fig:system_alfworld} to~\ref{fig:system_stabletoolbench}.
Ultimately, we collect 40K trajectories each for ALFWorld, SciWorld, and TextWorld, and 70K trajectories for WebShop on their respective training sets, as summarized in Table~\ref{tab:training_data_sizes}.

\begin{table}[!htb]
    \centering
    \resizebox{0.5\columnwidth}{!}{
    \begin{tabular}{lccc}
    \toprule
    Environment         & Train Games & Test Games & Trajectories \\
    \midrule
    ALFWorld           & 2420        & 200        & 40K                   \\
    SciWorld           & 2120        & 200        & 40K                   \\
    TextWorld          & 2500        & 200        & 40K                   \\
    WebShop            & 3930        & 200        & 70K                   \\
    StableToolBench    & 160K        & 2000         & None  \\
    \bottomrule
    \end{tabular}
    }
    \caption{Training data sizes for different environments. StableToolBench only contains single-turn training data without interactive trajectories.}
    \label{tab:training_data_sizes}
\end{table}

\paragraph{World Model Training Hyper-parameters} We utilize LLaMa-Facotry\footnote{\url{https://github.com/hiyouga/LLaMA-Factory}} \citep{zheng2024llamafactory} for SFT training of LLM-based world models. The training parameters are summarized in Table~\ref{tab:sft_training_parameters}. Parameters unspecified in the table follow the default settings of LLaMA-Factory.
Training data size varies across different environments, as detailed in Table~\ref{tab:training_data_sizes} if not otherwise specified.
Experiments are conducted on 4xH100-80GB GPUs.

\begin{table}[!htb]
    \centering
    \resizebox{0.5\columnwidth}{!}{
    \begin{tabular}{lc}
    \toprule
    Parameter                  & Value        \\
    \midrule
    Global Train Batch Size & 128            \\
    Learning Rate              & 1.0e-5       \\
    Number of Training Epochs  & 5            \\
    LR Scheduler Type          & Constant with Warmup \\
    Warmup Steps               & 10           \\
    BF16                       & True         \\
    Max Gradient Norm          & 100          \\
    \bottomrule
    \end{tabular}
    }
    \caption{SFT hyper-parameters for training LLM-based world models.}
    \label{tab:sft_training_parameters}
\end{table}

\paragraph{World Model Backbones} We utilize Qwen2.5-7B \citep{qwen2025qwen25technicalreport} and Llama3.1-8B \citep{grattafiori2024llama3herdmodels} as the primary backbone for LLM-based world models. To study the impact of model scale, we train Qwen2.5 models of four sizes: 0.5B, 1.5B, 3B, and 7B parameters. The specific model checkpoints used are as follows:

\begin{table}[!htb]
    \centering
    \resizebox{0.6\columnwidth}{!}{
    \begin{tabular}{ll}
    \toprule
    Model & Checkpoint URL \\
    \midrule
    Qwen2.5-7B & \url{https://huggingface.co/Qwen/Qwen2.5-7B} \\
    Qwen2.5-3B & \url{https://huggingface.co/Qwen/Qwen2.5-3B} \\
    Qwen2.5-1.5B & \url{https://huggingface.co/Qwen/Qwen2.5-1.5B} \\
    Qwen2.5-0.5B & \url{https://huggingface.co/Qwen/Qwen2.5-0.5B} \\
    Llama3.1-8B & \url{https://huggingface.co/meta-llama/Llama-3.1-8B} \\
    \bottomrule
    \end{tabular}
    }
    \caption{Model checkpoints used for world model training.}
    \label{tab:model_checkpoints}
\end{table}

\paragraph{API Models} We list the API models and their versions used in paper in Table~\ref{tab:llm_agent_versions}.

\begin{table}[!htb]
    \centering
    \resizebox{0.5\columnwidth}{!}{
    \begin{tabular}{ll}
    \toprule
    Model & Version \\
    \midrule
    GPT-4o-mini & gpt-4o-mini-2024-07-18 \\
    GPT-4o & gpt-4o-2024-11-20 \\
    GPT-4-turbo & gpt-4-turbo-2024-04-09 \\
    GPT-4.1 & gpt-4.1-2025-04-14 \\
    GPT-5 & gpt-5-2025-08-07 \\
    Gemini-2.5-flash & gemini-2.5-flash \\
    Claude-sonnet-4.5 & claude-sonnet-4-5-20250929 \\
    Qwen3-235B & qwen3-235b-a22b-instruct-2507 \\
    \bottomrule
    \end{tabular}
    }
    \caption{API models and versions used for evaluations.}
    \label{tab:llm_agent_versions}
\end{table}

\subsection{Synthetic Data Competes with Real}
\label{sec:appendix_syn_vs_real}
To compare the quality of world-model–synthesized trajectories with those from the real environment, we construct matched SFT datasets using 1,000 successful trajectories collected from (i) the real environment and (ii) the world model. To control for the behavior policy used during data collection, both datasets are generated by the same agent: a Qwen2.5-7B-Instruct policy trained via direct RL (i.e., without any SFT). This design avoids reusing the world-model training policy (GPT-4o) as the collector, thereby reducing the risk that the world model “self-replays” trajectories through the behavior policy. For world-model rollouts, trajectory success is determined by the model’s own predicted outcome.
\subsection{Early Experience for Policy Learning}
\label{sec:appendix_early_experience}

\paragraph{Early Experience (WM-SFT)}
To provide early dynamics exposure before policy learning, we warm-start the agent with a \emph{world-model style} supervised objective (Eq.~\ref{eq:worldmodel}): predicting the next environment response and termination signal conditioned on the dialogue history and the current action.
We use the same data sources described in Appendix~\ref{sec:appendix_implement} and sample 1{,}000 trajectories to construct the WM-SFT dataset.
Training follows the same SFT hyper-parameters as world model training (Table~\ref{tab:sft_training_parameters}).
For the baseline without early experience, this stage is skipped.

\paragraph{Agent Warmup (Agent-SFT)}
After WM-SFT, we perform a standard policy warmup stage by supervised fine-tuning on real-environment trajectories collected in Appendix~\ref{sec:appendix_implement}.
Specifically, we sample 1{,}000 trajectories and fine-tune the agent to generate its next turn (reasoning trace and action; Eq.~\ref{eq:agent}) from the interaction history.
We use the same SFT hyper-parameters as in Table~\ref{tab:sft_training_parameters}.

\paragraph{Reinforcement Learning (Agent-RL)} We utilize the AgentGymRL framework\footnote{\url{https://github.com/WooooDyy/AgentGym-RL}} \citep{xi2025agentgymrltrainingllmagents} to run GRPO training for agent policy training, with the suggested hyper-parameters as suggested in the paper. The command is as follows:

\begin{quote}
{\small
\begin{verbatim}
python3 -m verl.agent_trainer.main_ppo  \
    algorithm.adv_estimator=grpo \
    algorithm.rounds_ctrl.type=fixed \
    algorithm.rounds_ctrl.rounds=20 \
    data.train_file=${DATA_FILE} \
    data.train_batch_size=16 \
    data.max_prompt_length=1024 \
    data.max_response_length=4096 \
    actor_rollout_ref.agentgym.task_name=${TASK_NAME} \
    actor_rollout_ref.agentgym.env_addr=${ENV_ADDR} \
    actor_rollout_ref.agentgym.timeout=600 \
    actor_rollout_ref.model.path=${MODEL_PATH} \
    actor_rollout_ref.actor.use_kl_loss=True \
    actor_rollout_ref.actor.kl_loss_coef=0.001 \
    actor_rollout_ref.actor.kl_loss_type=low_var_kl \
    actor_rollout_ref.rollout.gpu_memory_utilization=0.6 \
    actor_rollout_ref.rollout.n=8 \
    actor_rollout_ref.rollout.max_model_len=32768 \
    actor_rollout_ref.rollout.max_tokens=200 \
    actor_rollout_ref.rollout.tensor_model_parallel_size=1 \
    actor_rollout_ref.actor.ppo_epochs=1 \
    actor_rollout_ref.actor.optim.lr=1e-6 \
    actor_rollout_ref.actor.ppo_mini_batch_size=8 \
    actor_rollout_ref.actor.ppo_micro_batch_size_per_gpu=1 \
    algorithm.kl_ctrl.kl_coef=0.001 \
    trainer.default_local_dir='outputs/ckpt/${trainer.experiment_name}' \
    trainer.project_name="agentgym" \
    trainer.experiment_name="${EXPERIMENT_NAME}" \
    trainer.save_freq=10 \
    trainer.total_epochs=10 \
    trainer.n_gpus_per_node=4
\end{verbatim}
}
\end{quote}

\subsection{World Model Initialization Context}\label{sec:appendix_world_model_initialization_context}
In \textit{ALFWorld} and \textit{SciWorld}, each game instance involves random initialization of the environment. For example, in \textit{ALFWorld}, the positions and contents of objects within rooms vary, while in \textit{SciWorld}, the connectivity of houses changes with each initialization. Consequently, even for humans, accurately predicting the next state of the environment based solely on task descriptions is challenging.
Similar to RAWM \citep{yang-etal-2025-efficient}, we include the initial state information of the environment for the world model's predictions. This design aligns with practical applications where the world model is used with knowledge of the initial environment state.
In data synthesis scenarios, such random states can be sampled through similar random generation methods, enhancing the diversity and generalization capabilities of the world model.
Examples of initial state information are provided in Figure~\ref{fig:init_state_alfworld} and~\ref{fig:init_state_sciworld}.
While \textit{TextWorld} lacks full initial state information due to environment limits. \textit{WebShop} and \textit{StableToolBench} are inherently open environments where comprehensive initial state information cannot be provided, so they also do not include initial state information.
\input{casestudy/init_state.tex}

\newpage
\section{Task Examples and Case Studies}\label{sec:appendix_task_examples_and_case_studies}
In this section, we provide task examples and case studies on world model across five environments.
\input{casestudy/stabletoolbench.tex}
\input{casestudy/alfworld.tex}
\input{casestudy/sciworld.tex}
\input{casestudy/textworld.tex}
\input{casestudy/webshop.tex}

\newpage
\section{Detailed Results}\label{sec:appendix_full_results}

We provide detailed results in Table~\ref{appendix:tab:ood_generalization} for the OOD generalization of world models.

\input{tables/alfworld_ood.tex}

\newpage
\section{System Prompts for Agent Trajectory Collection}\label{sec:appendix_system_prompts}
\input{casestudy/system_prompt.tex}

%% file: tables/env_description.tex
\begin{table*}[!ht]
    \centering
    \resizebox{\linewidth}{!}{
    \small
    \begin{tabular}{p{2.4cm} p{3.3cm} p{3.4cm} p{3.3cm} p{3.4cm}}
    \toprule
    \textbf{Environment} & \textbf{Description} & \textbf{Required Agent Ability} & \textbf{World Model State} & \textbf{Required World Model Ability} \\
    \midrule

    \textbf{ALFWorld} \citep{shridhar2021alfworldaligningtextembodied}
    & Embodied environment where agents accomplish household tasks by issuing text-based commands.
    & Spatial and physical commonsense, reasoning about containers and locations, and multi-step executions.
    & Room layout with hundreds of container–object combinations, agent inventory, and task progression.
    & Track physical configurations, maintain object relations, and predict stable multi-step state transitions.

    \\ \midrule

    \textbf{SciWorld} \citep{wang2022scienceworldagentsmarter5th}
    & Text-based interactivate laboratory environment involving simplified physics \& chemistry experiments.
    & Scientific concepts, causal reasoning, experiment planning, hypothesis testing with outcome evaluation.
    & Ten interconnected labs with \(\sim\)200 materials, intermediate substance states, and experiment progress.
    & Scientific dynamics modeling, physical reasoning, chemical simulation, experiment progress estimation.

    \\ \midrule

    \textbf{TextWorld} \citep{cote18textworld}
    & Text-based open-world environment supporting exploration, interaction, and diverse quest-like tasks.
    & Environment understanding, open-ended task planning, temporal tracking, and structured exploration.
    & Multiple connected rooms with \(\sim\)10 objects, exploration and discovery status, and task advancement.
    & Long-horizon state prediction, symbolic transition feedback, and exploration progress estimation.

    \\ \midrule

    \textbf{WebShop} \citep{yao2023webshopscalablerealworldweb}
    & Simulated shopping website where agents search, browse, and shop through multi-step interactions.
    & Goal decomposition, product evaluation, and robust reasoning over diverse semi-structured attributes.
    & Metadata for over 1M product attributes, search-query items surface, item details, and cart states.
    & Simulation of search engines, multi-step web navigation, product attributes, and constraint satisfaction.

    \\ \midrule

    \textbf{StableToolBench} \citep{guo2025stabletoolbenchstablelargescalebenchmarking}
    & API-based tool-use environment requiring schema adherence and structured output generation.
    & Doc understanding, symbolic reasoning, and executing schema-compliant action sequences.
    & Over 10K API tools, input/output schemas, intermediate tool-call states, and execution context.
    & Symbolic world state simulation, doc understanding, schema constraint satisfaction, structured generation.

    \\
    \bottomrule
    \end{tabular}
    }
    \caption{Summary of the five text-based environments used in our paper, highlighting the knowledge demands placed on both agents and world models. Task examples are provided in Figures \ref{fig:case_study_stabletoolbench}--\ref{fig:case_study_webshop} in Appendix \ref{sec:appendix_task_examples_and_case_studies}.}
    \label{tab:envs}
    \end{table*}

%% file: casestudy/init_state.tex
\begin{figure*}[htbp]
\centering
\resizebox{1.0\textwidth}{!}{  
\begin{tikzpicture}[
    node distance=0.15cm,
    every node/.style={align=left, font=\scriptsize}, 
    section/.style={draw, thick, minimum width=6.5cm, minimum height=5.2cm}, 
    header/.style={fill=gray!30, font=\bfseries\small, minimum width=6.5cm, minimum height=0.6cm},
    thought/.style={font=\bfseries},
    action/.style={font=\bfseries},
    observation/.style={font=\bfseries, color=blue!70},
    reward/.style={font=\bfseries, color=purple!70},
    highlight/.style={fill=yellow!50}
]

\node[draw, thick, minimum width=13.5cm, minimum height=1cm, fill=gray!10] (instruction) at (0, 0) {
    \begin{minipage}{13cm}
\# Environment Information (Only visible to Assistant)\\
\\
=== Objects on Receptacles ===\\
 - cabinet 1 is closed, if opened, in it, you see nothing.\\
 - cabinet 2 is closed, if opened, on the cabinet 2, you see a dishsponge 1\\
 - cabinet 3 is closed, if opened, on the cabinet 3, you see a mug 1\\
 - cabinet 4 is closed, if opened, in it, you see nothing.\\
 - cabinet 5 is closed, if opened, on the cabinet 5, you see a soapbottle 3\\
 - cabinet 6 is closed, if opened, on the cabinet 6, you see a plate 1\\
 - On the cabinet 9, you see a mug 3\\
 - On the countertop 1, you see a bread 1, a dishsponge 3, a egg 2, a fork 2, a mug 2, a pen 2, a peppershaker 1, a potato 1, a potato 2, a winebottle 2, a winebottle 3\\
 - On the diningtable 1, you see a apple 1, a bowl 1, a bowl 2, a butterknife 1, a butterknife 2, a butterknife 3, a fork 1, a glassbottle 1, a glassbottle 2, a knife 1, a knife 2, a lettuce 1, a pen 1, a pot 1, a saltshaker 1, a saltshaker 2, a soapbottle 1, a soapbottle 2, a spatula 1, a spatula 2, a winebottle 1\\
 - drawer 2 is closed, if opened, on the drawer 2, you see a dishsponge 2, a pencil 1\\
 - fridge 1 is closed, if opened, on the fridge 1, you see a bowl 3, a cup 1, a lettuce 2, a tomato 1\\
 - On the garbagecan 1, you see a apple 2, a egg 1\\
 - microwave 1 is closed, if opened, on the microwave 1, you see a apple 3\\
 - On the sinkbasin 1, you see a glassbottle 3, a spoon 1\\
 - On the stoveburner 1, you see a pan 1\\
 - On the stoveburner 3, you see a pan 2\\
 - On the stoveburner 4, you see a pan 2\\
\\
\# User Environment Information (Displayed to User)\\
\\
You are in the middle of a room. Looking quickly around you, you see a cabinet 10, a cabinet 9, a cabinet 8, a cabinet 7, a cabinet 6, a cabinet 5, a cabinet 4, a cabinet 3, a cabinet 2, a cabinet 1, a coffeemachine 1, a countertop 1, a diningtable 1, a drawer 2, a drawer 1, a fridge 1, a garbagecan 1, a microwave 1, a sinkbasin 1, a stoveburner 4, a stoveburner 3, a stoveburner 2, a stoveburner 1, and a toaster 1.\\
Your task is to: cool some mug and put it in coffeemachine.\\
AVAILABLE ACTIONS: go to cabinet 1,go to cabinet 10,go to cabinet 2,go to cabinet 3,go to cabinet 4,go to cabinet 5,go to cabinet 6,go to cabinet 7,go to cabinet 8,go to cabinet 9,go to coffeemachine 1,go to countertop 1,go to diningtable 1,go to drawer 1,go to drawer 2,go to fridge 1,go to garbagecan 1,go to microwave 1,go to sinkbasin 1,go to stoveburner 1,go to stoveburner 2,go to stoveburner 3,go to stoveburner 4,go to toaster 1,help,inventory,look
    \end{minipage}
};

\end{tikzpicture}
}  
\caption{Initialization Context Example of ALFWorld}
\label{fig:init_state_alfworld}
\end{figure*}

\begin{figure*}[htbp]
\centering
\resizebox{1.0\textwidth}{!}{  
\begin{tikzpicture}[
    node distance=0.15cm,
    every node/.style={align=left, font=\scriptsize}, 
    section/.style={draw, thick, minimum width=6.5cm, minimum height=5.2cm}, 
    header/.style={fill=gray!30, font=\bfseries\small, minimum width=6.5cm, minimum height=0.6cm},
    thought/.style={font=\bfseries},
    action/.style={font=\bfseries},
    observation/.style={font=\bfseries, color=blue!70},
    reward/.style={font=\bfseries, color=purple!70},
    highlight/.style={fill=yellow!50}
]

\node[draw, thick, minimum width=13.5cm, minimum height=1cm, fill=gray!10] (instruction) at (0, 0) {
    \begin{minipage}{13cm}
\# Environment Information (Only visible to Assistant)\\
\\
\\
=== Goal Progress ===\\
Completed keys: \\
----------------------------------------------------------------------------------------------------\\
Sequential Subgoals:\\
----------------------------------------------------------------------------------------------------\\
0       false                                   GoalFind        focus on thermometer\\
1       false                                   GoalFind        focus on substance\\
2       false                          GoalFindAnswerBox        focus on correct answer box\\
----------------------------------------------------------------------------------------------------\\
Unordered and Optional Subgoals:\\
----------------------------------------------------------------------------------------------------\\
0       false                       GoalInRoomWithObject        be in same location as thermometer\\
1       false        GoalSpecificObjectInDirectContainer        have thermometer in inventory\\
2       false                      GoalMoveToNewLocation        move to a new location\\
3       false                         GoalMoveToLocation        move to the location asked by the task (substance location)\\
4       false                         GoalMoveToLocation        move to the location asked by the task (answer box location)\\
5       false                         GoalMoveToLocation        move to a location with a heating device (kitchen)\\
6       false                         GoalMoveToLocation        move to a location with a heating device (outside)\\
7       false                         GoalMoveToLocation        move to a location with a heating device (foundry)\\
8       false        GoalSpecificObjectInDirectContainer        have task object in inventory\\
9       false            GoalPastActionUseObjectOnObject        use thermometer on substance\\
10      false            GoalPastActionUseObjectOnObject        use thermometer on substance (after it has been heated)\\
11      false               GoalObjectsInSingleContainer        have substance alone in a single container\\
12      false                 GoalActivateDeviceWithName        activate heater (stove)\\
13      false                 GoalActivateDeviceWithName        activate heater (blast furnace)\\
14      false                 GoalActivateDeviceWithName        activate heater (oven)\\
15      false                 GoalActivateDeviceWithName        activate heater (hot plate)\\
16      false        GoalSpecificObjectInDirectContainer        have lighter in inventory\\
17      false        GoalSpecificObjectInDirectContainer        move wood into fire pit\\
18      false                      GoalTemperatureOnFire        ignite wood\\
19      false                      GoalObjectInContainer        have substance on heater (stove)\\
20      false                      GoalObjectInContainer        have substance on heater (blast furnace)\\
21      false                      GoalObjectInContainer        have substance on heater (oven)\\
22      false                      GoalObjectInContainer        have substance on heater (hot plate)\\
23      false                      GoalObjectInContainer        have substance on heater (fire pit)\\
24      false                    GoalTemperatureIncrease        heat substance by at least 20C\\
----------------------------------------------------------------------------------------------------\\
\\
\\
=== Possible Actions ===\\
 - activate OBJ\\
 - close OBJ\\
 - connect OBJ to OBJ\\
 - deactivate OBJ\\
 - disconnect OBJ\\
 - dunk OBJ in OBJ\\
 - eat OBJ\\
 - flush OBJ\\
 - focus on OBJ\\
 - go OBJ\\
 - inventory\\
 - look around
    \end{minipage}
};

\end{tikzpicture}
}  
\caption{Initialization Context Example of SciWorld}
\label{fig:init_state_sciworld}
\end{figure*}

\begin{figure*}[htbp]
\centering
\resizebox{1.0\textwidth}{!}{  
\begin{tikzpicture}[
    node distance=0.15cm,
    every node/.style={align=left, font=\scriptsize}, 
    section/.style={draw, thick, minimum width=6.5cm, minimum height=5.2cm}, 
    header/.style={fill=gray!30, font=\bfseries\small, minimum width=6.5cm, minimum height=0.6cm},
    thought/.style={font=\bfseries},
    action/.style={font=\bfseries},
    observation/.style={font=\bfseries, color=blue!70},
    reward/.style={font=\bfseries, color=purple!70},
    highlight/.style={fill=yellow!50}
]

\node[draw, thick, minimum width=13.5cm, minimum height=1cm, fill=gray!10] (instruction) at (0, 0) {
    \begin{minipage}{13cm}

 - look at OBJ\\
 - look in OBJ\\
 - mix OBJ\\
 - move OBJ to OBJ\\
 - open OBJ\\
 - pick up OBJ\\
 - pour OBJ in OBJ\\
 - put down OBJ\\
 - read OBJ\\
 - reset task\\
 - task\\
 - teleport OBJ\\
 - use OBJ on OBJ\\
 - wait\\
 - wait1\\
\\

=== Per-Room Observations ===\\
== Room: terminal 2 ==\\
This room is called the bathroom. In it, you see: \\
        the agent\\
        a substance called air\\
        a bathtub, which is turned off. In the bathtub is: nothing.\\
        a glass cup (containing nothing)\\
        a picture\\
        a sink, which is turned off. In the sink is: nothing.\\
        a toilet. In the toilet is: A drain, which is open, a substance called water.\\
You also see:\\
        A door to the kitchen (that is closed)\\
Possible Objects: agent, air, bathroom, bathtub, cup, door, kitchen, picture, sink, substance in toilet, toilet\\
\\
== Room: kitchen ==\\
This room is called the bathroom. In it, you see: \\
        the agent\\
        a substance called air\\
        a bathtub, which is turned off. In the bathtub is: nothing.\\
        a glass cup (containing nothing)\\
        a picture\\
        a sink, which is turned off. In the sink is: nothing.\\
        a toilet. In the toilet is: A drain, which is closed, a substance called water.\\
You also see:\\
        A door to the kitchen (that is closed)\\
Possible Objects: agent, air, bathroom, bathtub, cup, door, kitchen, picture, sink, substance in toilet, toilet\\
\\
== Room: sewer ==\\
This room is called the bathroom. In it, you see: \\
        the agent\\
        a substance called air\\
        a bathtub, which is turned off. In the bathtub is: nothing.\\
        a glass cup (containing nothing)\\
        a picture\\
        a sink, which is turned off. In the sink is: nothing.\\
        a toilet. In the toilet is: A drain, which is closed, a substance called water.\\
You also see:\\
        A door to the kitchen (that is closed)\\
Possible Objects: agent, air, bathroom, bathtub, cup, door, kitchen, picture, sink, substance in toilet, toilet
    \end{minipage}
};

\end{tikzpicture}
}  
\caption{Initialization Context Example of SciWorld (Continued)}
\label{fig:init_state_sciworld_cont1}
\end{figure*}

\begin{figure*}[htbp]
\centering
\resizebox{1.0\textwidth}{!}{  
\begin{tikzpicture}[
    node distance=0.15cm,
    every node/.style={align=left, font=\scriptsize}, 
    section/.style={draw, thick, minimum width=6.5cm, minimum height=5.2cm}, 
    header/.style={fill=gray!30, font=\bfseries\small, minimum width=6.5cm, minimum height=0.6cm},
    thought/.style={font=\bfseries},
    action/.style={font=\bfseries},
    observation/.style={font=\bfseries, color=blue!70},
    reward/.style={font=\bfseries, color=purple!70},
    highlight/.style={fill=yellow!50}
]

\node[draw, thick, minimum width=13.5cm, minimum height=1cm, fill=gray!10] (instruction) at (0, 0) {
    \begin{minipage}{13cm}
== Room: bedroom ==\\
This room is called the bedroom. In it, you see: \\
        the agent\\
        a substance called air\\
        a bed. On the bed is: a mattress. On the mattress is: a white pillow..\\
        a book shelf (containing nothing)\\
        a closet. The closet door is closed. \\
        a drawing\\
        a table. On the table is: nothing.\\
You also see:\\
        A door to the hallway (that is closed)\\
Possible Objects: agent, air, bed, bedroom, book shelf, closet, cloth sittable, door, drawing, hallway, object, table\\
\\
== Room: art studio ==\\
This room is called the art studio. In it, you see: \\
        the agent\\
        a substance called air\\
        a large cupboard. The large cupboard door is closed. \\
        a table. On the table is: a jug (containing nothing).\\
        a wood cup (containing yellow paint)\\
        a wood cup (containing blue paint)\\
        a wood cup (containing red paint)\\
You also see:\\
        A door to the hallway (that is closed)\\
Possible Objects: agent, air, art studio, blue paint, cup containing blue paint, cup containing red paint, cup containing yellow paint, cupboard, door, hallway, jug, paint in cup containing red paint, paint in cup containing yellow paint, table\\
\\
== Room: foundry ==\\
This room is called the foundry. In it, you see: \\
        the agent\\
        a substance called air\\
        a blast furnace, which is turned off. The blast furnace door is closed. \\
        a sink, which is turned off. In the sink is: nothing.\\
        a table. On the table is: nothing.\\

You also see:\\
        A door to the outside (that is closed)\\
Possible Objects: agent, air, blast furnace, door, foundry, outside, sink, steel table\\
\\
== Room: workshop ==\\
This room is called the workshop. In it, you see: \\
        the agent\\
        a substance called air\\
        a table. On the table is: a battery, a black wire, a blue light bulb, which is off, a green light bulb, which is off, a red wire, a switch, which is off, a violet light bulb, which is off, a yellow wire.\\
        a ultra low temperature freezer. The ultra low temperature freezer door is closed. \\
You also see:\\
        A door to the hallway (that is closed)\\
Possible Objects: agent, air, anode in battery, anode in blue light bulb, anode in component, anode in green light bulb, anode in violet light bulb, battery, battery cathode, black wire, black wire terminal 1, black wire terminal 2, blue light bulb, blue light bulb cathode, cathode in component, cathode in green light bulb, cathode in violet light bulb, door, freezer, green light bulb, hallway, red wire, red wire terminal 1, red wire terminal 2, switch, table, terminal 1 in yellow wire, terminal 2 in yellow wire, violet light bulb, workshop, yellow wire
    \end{minipage}
};

\end{tikzpicture}
}  
\caption{Initialization Context Example of SciWorld (Continued)}
\label{fig:init_state_sciworld_cont2}
\end{figure*}

\begin{figure*}[htbp]
\centering
\resizebox{1.0\textwidth}{!}{  
\begin{tikzpicture}[
    node distance=0.15cm,
    every node/.style={align=left, font=\scriptsize}, 
    section/.style={draw, thick, minimum width=6.5cm, minimum height=5.2cm}, 
    header/.style={fill=gray!30, font=\bfseries\small, minimum width=6.5cm, minimum height=0.6cm},
    thought/.style={font=\bfseries},
    action/.style={font=\bfseries},
    observation/.style={font=\bfseries, color=blue!70},
    reward/.style={font=\bfseries, color=purple!70},
    highlight/.style={fill=yellow!50}
]

\node[draw, thick, minimum width=13.5cm, minimum height=1cm, fill=gray!10] (instruction) at (0, 0) {
    \begin{minipage}{13cm}
== Room: hallway ==\\
This room is called the hallway. In it, you see: \\
        the agent\\
        a substance called air\\
        a drawing\\
You also see:\\
        A door to the art studio (that is closed)\\
        A door to the bedroom (that is closed)\\
        A door to the greenhouse (that is closed)\\
        A door to the kitchen (that is closed)\\
        A door to the living room (that is closed)\\
        A door to the workshop (that is closed)\\
Possible Objects: agent, air, art studio, art studio door, bedroom, bedroom door, door to greenhouse, door to kitchen, door to living room, door to workshop, drawing, greenhouse, hallway, kitchen, living room, workshop\\
\\
== Room: bathroom ==\\
This room is called the bathroom. In it, you see: \\
        the agent\\
        a substance called air\\
        a bathtub, which is turned off. In the bathtub is: nothing.\\
        a glass cup (containing nothing)\\
        a picture\\
        a sink, which is turned off. In the sink is: nothing.\\
        a toilet. In the toilet is: A drain, which is closed, a substance called water.\\
You also see:\\
        A door to the kitchen (that is closed)\\
Possible Objects: agent, air, bathroom, bathtub, cup, door, kitchen, picture, sink, substance in toilet, toilet\\
\\
== Room: terminal 1 ==\\
This room is called the bathroom. In it, you see: \\
        the agent\\
        a substance called air\\
        a bathtub, which is turned off. In the bathtub is: nothing.\\
        a glass cup (containing nothing)\\
        a picture\\
        a sink, which is turned off. In the sink is: nothing.\\
        a toilet. In the toilet is: A drain, which is closed, a substance called water.\\
You also see:\\
        A door to the kitchen (that is closed)\\
Possible Objects: agent, air, bathroom, bathtub, cup, door, kitchen, picture, sink, substance in toilet, toilet\\
\\
== Room: living room ==\\
This room is called the bathroom. In it, you see: \\
        the agent\\
        a substance called air\\
        a bathtub, which is turned off. In the bathtub is: nothing.\\
        a glass cup (containing nothing)\\
        a picture\\
        a sink, which is turned off. In the sink is: nothing.\\
        a toilet. In the toilet is: A drain, which is closed, a substance called water.\\
You also see:\\
        A door to the kitchen (that is closed)\\
Possible Objects: agent, air, bathroom, bathtub, cup, door, kitchen, picture, sink, substance in toilet, toilet
    \end{minipage}
};

\end{tikzpicture}
}  
\caption{Initialization Context Example of SciWorld (Continued)}
\label{fig:init_state_sciworld_cont3}
\end{figure*}

\begin{figure*}[htbp]
\centering
\resizebox{1.0\textwidth}{!}{  
\begin{tikzpicture}[
    node distance=0.15cm,
    every node/.style={align=left, font=\scriptsize}, 
    section/.style={draw, thick, minimum width=6.5cm, minimum height=5.2cm}, 
    header/.style={fill=gray!30, font=\bfseries\small, minimum width=6.5cm, minimum height=0.6cm},
    thought/.style={font=\bfseries},
    action/.style={font=\bfseries},
    observation/.style={font=\bfseries, color=blue!70},
    reward/.style={font=\bfseries, color=purple!70},
    highlight/.style={fill=yellow!50}
]

\node[draw, thick, minimum width=13.5cm, minimum height=1cm, fill=gray!10] (instruction) at (0, 0) {
    \begin{minipage}{13cm}
== Room: outside ==\\
This outside location is called the outside. Here you see: \\
        the agent\\
        a substance called air\\
        an axe\\
        a fire pit (containing nothing)\\
        a fountain (containing a substance called water)\\
        the ground\\
        a shovel\\
        a substance called wood\\
You also see:\\
        A door to the foundry (that is closed)\\
        A door to the greenhouse (that is closed)\\
        A door to the kitchen (that is closed)\\
Possible Objects: agent, air, axe, door to foundry, door to greenhouse, door to kitchen, fire pit, foundry, fountain, greenhouse, ground, kitchen, outside, shovel, substance in fountain, wood\\
\\
== Room: greenhouse ==\\
This room is called the greenhouse. In it, you see: \\
        the agent\\
        a substance called air\\
        a bee hive. The bee hive door is closed. \\
        a jug (containing nothing)\\
        a sink, which is turned off. In the sink is: nothing.\\
You also see:\\
        A door to the hallway (that is closed)\\
        A door to the outside (that is closed)\\
Possible Objects: agent, air, bee hive, door to hallway, door to outside, greenhouse, hallway, jug, outside, sink\\
\\
\\
\\
\# User Environment Information (Displayed to User)\\
\\
Your task is to measure the melting point of tin, which is located around the kitchen. First, focus on the thermometer. Next, focus on the tin. If the melting point of tin is above 200.0 degrees celsius, focus on the blue box. If the melting point of tin is below 200.0 degrees celsius, focus on the orange box. The boxes are located around the kitchen.\\
This room is called the bathroom. In it, you see: \\
        the agent\\
        a substance called air\\
        a bathtub, which is turned off. In the bathtub is: nothing.\\
        a glass cup (containing nothing)\\
        a picture\\
        a sink, which is turned off. In the sink is: nothing.\\
        a toilet. In the toilet is: A drain, which is open, a substance called water.\\
You also see:\\
        A door to the kitchen (that is closed)
    \end{minipage}
};

\end{tikzpicture}
}  
\caption{Initialization Context Example of SciWorld (Continued)}
\label{fig:init_state_sciworld_cont4}
\end{figure*}

%% file: casestudy/stabletoolbench.tex
\begin{figure*}[htbp]
\centering
\resizebox{1.0\textwidth}{!}{  
\begin{tikzpicture}[
    node distance=0.15cm,
    every node/.style={align=left, font=\tiny}, 
    section/.style={draw, thick, minimum width=6.5cm, minimum height=5cm}, 
    header/.style={fill=gray!30, font=\bfseries\small, minimum width=6.5cm, minimum height=0.6cm},
    thought/.style={font=\bfseries},
    action/.style={font=\bfseries},
    observation/.style={font=\bfseries, color=blue!70},
    reward/.style={font=\bfseries, color=purple!70},
    highlight/.style={fill=yellow!50}
]

\node[draw, thick, minimum width=13.5cm, minimum height=1cm, fill=gray!10] (instruction) at (0, 0) {
    \begin{minipage}{13cm}
    \textbf{Instruction}: API doc:\\
\{\\
``api\_name'': ``SWIFT codes by country'',\\
``api\_description'': ``Returns list of SWIFT codes for a country (use two letter ISO country code)'',\\
``required\_parameters'': [\\
\{ ``name'': ``country\_code'', ``type'': ``string'', ``description'': ``, ``default'': ``AU'' \}\\
],\\
``optional\_parameters'': [\,],\\
``tool\_description'': ``Lookup country banks and their SWIFT codes'',\\
``tool\_name'': ``Bank SWIFT codes'',\\
``tool\_category'': ``Financial''\\
\}\\
\\
Request:\\
\{\\
``category'': ``Financial'',\\
``tool\_name'': ``Bank SWIFT codes'',\\
``api\_name'': ``SWIFT codes by country'',\\
``tool\_input'': ``\{``country\_code'': ``JP''\}'',\\
``strip'': ``filter''\\
\}

    \end{minipage}
};

\node[section, fill=lightpink, below=0.5cm of instruction.south west, anchor=north west] (before) {};
\node[header, below=0.5cm of instruction.south west, anchor=north west] (before_header) {Real Trajectory};

\node[below=0.2cm of before_header.south west, anchor=north west, text width=6cm] (before_content) {\{\\
``error'': ``'',\\
``response'': ``\{ `data': [\\
\{ `country': `JP', `code': `FCTOJPJ1XXX', `bank': `ABN AMRO CLEARING TOKYO CO., LTD.' \},\\
\{ `country': `JP', `code': `ACSRJPJ1XXX', `bank': `ACE SECURITIES CO., LTD.' \},\\
\{ `country': `JP', `code': `ADCTJPJJXXX', `bank': `ADEKA CORPORATION' \},\\
\{ `country': `JP', `code': `AEONJPJ1XXX', `bank': `AEON BANK, LTD.' \},\\
\{ `country': `JP', `code': `AEONJPJTXXX', `bank': `AEON BANK, LTD.' \}...\\
] \}''\\
\}

};

\node[section, fill=lightblue, below=0.5cm of instruction.south east, anchor=north east] (after) {};
\node[header, below=0.5cm of instruction.south east, anchor=north east] (after_header) {World Model Trajectory};

\node[below=0.2cm of after_header.south west, anchor=north west, text width=6cm] (after_content) {\{\\
``error'': ``'',\\
``response'': ``\{ `data': [\\
\{ `country': `JP', `code': `01234567', `bank': `MUKAMAL BANK' \},\\
\{ `country': `JP', `code': `09876543', `bank': `NIPPON BANK' \},\\
\{ `country': `JP', `code': `12345678', `bank': `YAMATO BANK' \},\\
\{ `country': `JP', `code': `87654321', `bank': `TOHOKU BANK' \}\\
] \}''\\
\}
};

\end{tikzpicture}
}  
\caption{Task Example and Case Study of StableToolBench}
\label{fig:case_study_stabletoolbench}
\end{figure*}

%% file: casestudy/alfworld.tex
\begin{figure*}[htbp]
\centering
\resizebox{1.0\textwidth}{!}{  
\begin{tikzpicture}[
    node distance=0.15cm,
    every node/.style={align=left, font=\tiny}, 
    section/.style={draw, thick, minimum width=6.5cm, minimum height=14.5cm}, 
    header/.style={fill=gray!30, font=\bfseries\small, minimum width=6.5cm, minimum height=0.6cm},
    thought/.style={font=\bfseries},
    action/.style={font=\bfseries},
    observation/.style={font=\bfseries, color=blue!70},
    reward/.style={font=\bfseries, color=purple!70},
    highlight/.style={fill=yellow!50}
]

\node[draw, thick, minimum width=13.5cm, minimum height=1cm, fill=gray!10] (instruction) at (0, 0) {
    \begin{minipage}{13cm}
    \textbf{Instruction}: You are in the middle of a room. Looking quickly around you, you see a armchair 1, a bed 1, a diningtable 1, a drawer 2, a drawer 1, a garbagecan 1, a sidetable 2, and a sidetable 1.\\Your task is to: put some pillow on armchair.\\AVAILABLE ACTIONS: go to armchair 1,go to bed 1,go to diningtable 1,go to drawer 1,go to drawer 2,go to garbagecan 1,go to sidetable 1,go to sidetable 2,help,inventory,look
    \end{minipage}
};

\node[section, fill=lightpink, below=0.5cm of instruction.south west, anchor=north west] (before) {};
\node[header, below=0.5cm of instruction.south west, anchor=north west] (before_header) {Real Trajectory};

\node[below=0.2cm of before_header.south west, anchor=north west, text width=6cm] (before_content) {
    \textbf{Thought 1:}
    My task is to put a pillow on the armchair. Before proceeding, I should first find out where the pillows are located. A likely place could be the bed or the sidetables. Starting with checking the bed might be a good idea as pillows are commonly found there.\\[0.1cm]

    \textbf{Action 1:}
    go to bed 1\\[0.1cm]

    \color{blue!70}\textbf{Observation 1:}
    \color{black}You arrive at bed 1. On the bed 1, you see a book 2, a laptop 1, a pillow 2, and a pillow 1.\\[0.15cm]

    \textbf{Thought 2:}
    I have found two pillows on the bed. My task requires putting a pillow on the armchair. I need to pick up one of the pillows and then go to the armchair.\\[0.1cm]

    \textbf{Action 2:}
    inventory\\[0.1cm]

    \color{blue!70}\textbf{Observation 2:}
    \color{black}You are not carrying anything.\\[0.15cm]

    \textbf{Thought 3:}
    I need to pick up a pillow from the bed to carry it to the armchair. Once I have the pillow, I'll proceed to go to the armchair to complete the task.\\[0.1cm]

    \textbf{Action 3:}
    pick up pillow 1\\[0.1cm]

    \color{blue!70}\textbf{Observation 3:}
    \color{black}Nothing happens.\\[0.15cm]

    \textbf{Thought 4:}
    The action to pick up the pillow didn't work, suggesting perhaps it's not a valid action at this moment, or I need to utilize another approach to be able to pick up items. I should consult the help to see if there's any guidance on how to interact with objects in the environment.\\[0.1cm]

    \textbf{Action 4:}
    help\\[0.1cm]

    \color{blue!70}\textbf{Observation 4:}
    \color{black}Available commands: look: look around your current location.  inventory: check your current inventory.  go to (receptacle):  move to a receptacle.  open (receptacle):  open a receptacle...\\[0.15cm]

    \textbf{Thought 5:}
    Based on the available commands, I should use the "take (object) from (receptacle)" command to pick up a pillow from the bed. I will proceed with the correct format to take a pillow.\\[0.1cm]

    \textbf{Action 5:}
    take pillow 1 from bed 1\\[0.1cm]

    \color{blue!70}\textbf{Observation 5:}
    \color{black}You pick up the pillow 1 from the bed 1.\\[0.15cm]

    \textbf{Thought 6:}
    I successfully picked up a pillow. Now, I need to go to the armchair to place the pillow on it, completing the task.\\[0.1cm]

    \textbf{Action 6:}
    go to armchair 1\\[0.1cm]

    \color{blue!70}\textbf{Observation 6:}
    \color{black}You arrive at armchair 1. On the armchair 1, you see nothing.\\[0.15cm]

    \textbf{Thought 7:}
    I'm at the armchair where I need to place the pillow. I will use the \"move (object) to (receptacle)\" command to put the pillow on the armchair.\\[0.1cm]

    \textbf{Action 7:}
    move pillow 1 to armchair 1\\[0.1cm]

    \color{blue!70}\textbf{Observation 7:}
    \color{black}You move the pillow 1 to the armchair 1.\\[0.15cm]

    \color{purple!70}\textbf{Success:} \color{black}True
};

\node[section, fill=lightblue, below=0.5cm of instruction.south east, anchor=north east] (after) {};
\node[header, below=0.5cm of instruction.south east, anchor=north east] (after_header) {World Model Trajectory};

\node[below=0.2cm of after_header.south west, anchor=north west, text width=6cm] (after_content) {
    \textbf{Thought 1:}
    In order to complete the task of putting a pillow on the armchair, I need to first find a pillow. Pillows are most likely to be found on a bed, in a drawer, or on a sidetable. I should explore these areas to locate a pillow and then proceed to the armchair to place it there.\\[0.1cm]

    \textbf{Action 1:}
    go to bed 1\\[0.1cm]

    \color{blue!70}\textbf{Observation 1:}
    \color{black}You arrive at bed 1. On the bed 1, you see a book 2, a laptop 1, a pillow 2, and a pillow 1.\\[0.15cm]

    \textbf{Thought 2:}
    I have found pillows on the bed, which is helpful in achieving the task. I need to pick up one of the pillows so I can then place it on the armchair.\\[0.1cm]

    \textbf{Action 2:}
    inventory\\[0.1cm]

    \color{blue!70}\textbf{Observation 2:}
    \color{black}You are not carrying anything.\\[0.15cm]

    \textbf{Thought 3:}
    It's confirmed that I am not carrying anything. To complete the task, I should pick up one of the pillows from the bed and then make my way to the armchair.\\[0.1cm]

    \textbf{Action 3:}
    Take pillow 1\\[0.1cm]

    \color{blue!70}\textbf{Observation 3:}
    \color{black}Nothing happens.\\[0.15cm]

    \textbf{Thought 4:}
    It appears that "Take pillow 1" is not a valid action, or perhaps it is not the correct format for the given scenario. I will revise my approach and check if there is a specific action for picking up items or if I need to use a different method.\\[0.1cm]

    \textbf{Action 4:}
    help\\[0.1cm]

    \color{blue!70}\textbf{Observation 4:}
    \color{black}Available commands: look: look around your current location.  inventory: check your current inventory.  go to (receptacle):  move to a receptacle.  open (receptacle):  open a receptacle...\\[0.15cm]

    \textbf{Thought 5:}
    Now that I have the list of available commands, I see that I need to use the "take (object) from (receptacle)" command to pick up a pillow from the bed. I will proceed to take one of the pillows.\\[0.1cm]

    \textbf{Action 5:}
    take pillow 1 from bed 1\\[0.1cm]

    \color{blue!70}\textbf{Observation 5:}
    \color{black}You pick up the pillow 1 from the bed 1.\\[0.15cm]

    \textbf{Thought 6:}
    I now have pillow 1 in my possession. The next step is to go to the armchair so that I can place the pillow on it to achieve the task goal.\\[0.1cm]

    \textbf{Action 6:}
    go to armchair 1\\[0.1cm]

    \color{blue!70}\textbf{Observation 6:}
    \color{black}You arrive at armchair 1. On the armchair 1, you see nothing.\\[0.15cm]

    \textbf{Thought 7:}
    I am now at the armchair with pillow 1 in my inventory. The final step is to place the pillow onto the armchair to complete the task.\\[0.1cm]

    \textbf{Action 7:}
    move pillow 1 to armchair 1\\[0.1cm]

    \color{blue!70}\textbf{Observation 7:}
    \color{black}You move the pillow 1 to the armchair 1\\[0.15cm]




    \color{purple!70}\textbf{Success:} \color{black}True
};

\end{tikzpicture}
}  
\caption{Task Example and Case Study of ALFWorld}
\label{fig:case_study_alfworld} 
\end{figure*}

%% file: casestudy/sciworld.tex
\begin{figure*}[htbp]
\centering
\resizebox{1.0\textwidth}{!}{  
\begin{tikzpicture}[
    node distance=0.15cm,
    every node/.style={align=left, font=\tiny}, 
    section/.style={draw, thick, minimum width=6.5cm, minimum height=14.5cm}, 
    header/.style={fill=gray!30, font=\bfseries\small, minimum width=6.5cm, minimum height=0.6cm},
    thought/.style={font=\bfseries},
    action/.style={font=\bfseries},
    observation/.style={font=\bfseries, color=blue!70},
    reward/.style={font=\bfseries, color=purple!70},
    highlight/.style={fill=yellow!50}
]

\node[draw, thick, minimum width=13.5cm, minimum height=1cm, fill=gray!10] (instruction) at (0, 0) {
    \begin{minipage}{13cm}
    \textbf{Instruction}: Your task is to measure the temperature of unknown substance B, which is located around the bathroom. First, focus on the thermometer. Next, focus on the unknown substance B. If the unknown substance B temperature is above 0.0 degrees celsius, place it in the yellow box. If the unknown substance B temperature is below 0.0 degrees celsius, place it in the purple box. The boxes are located around the bathroom.\\ This room is called the workshop. In it, you see: \\ the agent\\ a substance called air\\ a table. On the table is: a battery, a blue light bulb, which is off, a blue wire, a orange wire, a red wire, a switch, which is off, a violet light bulb, which is off, a yellow light bulb, which is off.\\ a ultra low temperature freezer. The ultra low temperature freezer door is closed. \\ You also see: A door to the hallway (that is closed)
    \end{minipage}
};

\node[section, fill=lightpink, below=0.5cm of instruction.south west, anchor=north west] (before) {};
\node[header, below=0.5cm of instruction.south west, anchor=north west] (before_header) {Real Trajectory};

\node[below=0.2cm of before_header.south west, anchor=north west, text width=6cm] (before_content) {
    \textbf{Thought 1:}
    To start this task, I first need to focus on the thermometer, as it is crucial for measuring the temperature of the unknown substance B.\\[0.1cm]

    \textbf{Action 1:}
    focus on thermometer\\[0.1cm]

    \color{blue!70}\textbf{Observation 1:}
    \color{black}No known action matches that input.\\[0.15cm]

    \textbf{Thought 2:}
    It seems I need to revise my approach. Let me start by looking around to gather more information about my surroundings and locate the thermometer.\\[0.1cm]

    \textbf{Action 2:}
    look around\\[0.1cm]

    \color{blue!70}\textbf{Observation 2:}
    \color{black}This room is called the workshop. In it, you see: ...\\[0.15cm]

    \textbf{Thought 3:}
    I don't see any mention of a thermometer in my surroundings. I should explore further to find where the thermometer might be located or check the areas close to the bathroom for their contents.\\[0.1cm]

    \textbf{Action 3:}
    go to the hallway\\[0.1cm]

    \color{blue!70}\textbf{Observation 3:}
    \color{black}The door is not open.\\[0.15cm]

    \textbf{Thought 4:}
    I need to open the door to the hallway so I can proceed to search for the thermometer and unknown substance B near the bathroom.\\[0.1cm]

    \textbf{Action 4:}
    open door to the hallway\\[0.1cm]

    \color{blue!70}\textbf{Observation 4:}
    \color{black}The door is now open.\\[0.15cm]

    \textbf{Thought 5:}
    Now that the door to the hallway is open, I can proceed to the hallway and then look for the bathroom where the thermometer and unknown substance B might be.\\[0.1cm]

    \textbf{Action 5:}
    go to hallway\\[0.1cm]

    \color{blue!70}\textbf{Observation 5:}
    \color{black}You move to the hallway.\\[0.15cm]

    \textbf{Thought 6:}
    I am now in the hallway. I need to locate the bathroom to look for the thermometer and the unknown substance B.\\[0.1cm]

    \textbf{Action 6:}
    look around\\[0.1cm]

    \color{blue!70}\textbf{Observation 6:}
    \color{black}This room is called the hallway. In it, you see: \\ the agent\\ a substance called air\\ a finger painting\\You also see: A door to the art studio (that is closed)...\\[0.15cm]

    \textbf{[More steps omitted for brevity]}\\ [0.2cm]

    \textbf{Thought 24:}
    I've focused on the unknown substance B and previously focused on the thermometer. Now I ...\\[0.1cm]

    \textbf{Action 24:}
    move unknown substance B to yellow box\\[0.1cm]

    \color{blue!70}\textbf{Observation 24:}
    \color{black}You move the unknown substance B to the yellow box.\\[0.15cm]




    \color{purple!70}\textbf{Success:} \color{black}True
};

\node[section, fill=lightblue, below=0.5cm of instruction.south east, anchor=north east] (after) {};
\node[header, below=0.5cm of instruction.south east, anchor=north east] (after_header) {World Model Trajectory};

\node[below=0.2cm of after_header.south west, anchor=north west, text width=6cm] (after_content) {
    \textbf{Thought 1:}
    I need to find a thermometer to measure the temperature of the unknown substance B. Currently, I don't see a thermometer in the workshop, so I'll need to look around or go to a different room to find it.\\[0.1cm]

    \textbf{Action 1:}
    look around\\[0.1cm]

    \color{blue!70}\textbf{Observation 1:}
    \color{black}This room is called the workshop. In it, you see: ...\\[0.15cm]

    \textbf{Thought 2:}
    I still don't see a thermometer. I need to search more thoroughly or consider moving to another location. The door to the hallway could lead to another room where the thermometer or the unknown substance B might be found.\\[0.1cm]

    \textbf{Action 2:}
    open the door to the hallway\\[0.1cm]

    \color{blue!70}\textbf{Observation 2:}
    \color{black}The door is now open.\\[0.15cm]

    \textbf{Thought 3:}
    With the door to the hallway open, I can explore further to potentially locate the thermometer or the unknown substance B needed for the task.\\[0.1cm]

    \textbf{Action 3:}
    go to hallway\\[0.1cm]

    \color{blue!70}\textbf{Observation 3:}
    \color{black}You move to the hallway.\\[0.15cm]

    \textbf{Thought 4:}
    Now that I'm in the hallway, I should look around to see if there are any signs of the thermometer or the unknown substance B.\\[0.1cm]

    \textbf{Action 4:}
    look around\\[0.1cm]

    \color{blue!70}\textbf{Observation 4:}
    \color{black}This room is called the hallway. In it, you see: \\ the agent\\ a substance called air\\ a finger painting\\You also see: A door to the art studio (that is closed)...\\[0.15cm]

    \textbf{Thought 5:}
    There are several doors leading to different rooms. I should choose a room that is likely to ...\\[0.1cm]

    \textbf{Action 5:}
    go to the kitchen\\[0.1cm]

    \color{blue!70}\textbf{Observation 5:}
    \color{black}The door is not open.\\[0.15cm]

    \textbf{Thought 6:}
    I need to open the door to the kitchen to proceed and see if the thermometer is there.\\[0.1cm]

    \textbf{Action 6:}
    open the door to the kitchen\\[0.1cm]

    \color{blue!70}\textbf{Observation 6:}
    \color{black}The door is now open.\\[0.15cm]




    \textbf{[More steps omitted for brevity]}\\ [0.2cm]

    \textbf{Thought 20:}
    It seems I'm unable to use the actions incorrectly to check the temperature of the unknown ...\\[0.1cm]

    \textbf{Action 20:}
    move unknown substance B to yellow box\\[0.1cm]

    \color{blue!70}\textbf{Observation 20:}
    \color{black}You move the unknown substance B to the yellow box.\\[0.15cm]

    \color{purple!70}\textbf{Success:} \color{black}True
};

\end{tikzpicture}
}  
\caption{Task Example and Case Study of SciWorld}
\label{fig:case_study_sciworld} 
\end{figure*}

%% file: casestudy/textworld.tex
\begin{figure*}[htbp]
\centering
\resizebox{1.0\textwidth}{!}{  
\begin{tikzpicture}[
    node distance=0.15cm,
    every node/.style={align=left, font=\tiny}, 
    section/.style={draw, thick, minimum width=6.5cm, minimum height=15cm}, 
    header/.style={fill=gray!30, font=\bfseries\small, minimum width=6.5cm, minimum height=0.6cm},
    thought/.style={font=\bfseries},
    action/.style={font=\bfseries},
    observation/.style={font=\bfseries, color=blue!70},
    reward/.style={font=\bfseries, color=purple!70},
    highlight/.style={fill=yellow!50}
]

\node[draw, thick, minimum width=13.5cm, minimum height=1cm, fill=gray!10] (instruction) at (0, 0) {
    \begin{minipage}{13cm}
    \textbf{Instruction}: Hey, thanks for coming over to the TextWorld today, there is something I need you to do for me. First of all, you could, like, look and see that the antique trunk inside the bedroom is opened. Then, recover the old key from the antique trunk. Then, make absolutely sure that the wooden door inside the bedroom is unlocked. After unlocking the wooden door, open the wooden door in the bedroom. Then, try to head east. After that, try to travel south. Once you get through with that, take the milk from the couch within the living room. Having taken the milk, attempt to travel north. That done, rest the milk on the stove inside the kitchen. And if you do that, you're the winner!\\-= Bedroom =-\\You make a grand eccentric entrance into a bedroom.You see a closed normal looking chest drawer right there by you. You can see an antique trunk. You see a king-size bed. But the thing is empty.There is a closed wooden door leading east.\\-= Bedroom =-0/1\\AVAILABLE ACTIONS: examine antique trunk, examine chest drawer, examine king-size bed, examine wooden door, inventory, look, open antique trunk, open chest drawer
    \end{minipage}
};

\node[section, fill=lightpink, below=0.5cm of instruction.south west, anchor=north west] (before) {};
\node[header, below=0.5cm of instruction.south west, anchor=north west] (before_header) {Real Trajectory};

\node[below=0.2cm of before_header.south west, anchor=north west, text width=6cm] (before_content) {
    \textbf{Thought 1:}
    The description mentioned an antique trunk inside the bedroom that needs to be opened first. Since one of the admissible actions is to open the antique trunk, this should be the first step to proceed with the tasks.\\[0.1cm]

    \textbf{Action 1:}
    open antique trunk\\[0.1cm]

    \color{blue!70}\textbf{Observation 1:}
    \color{black}You open the antique trunk, revealing an old key.\\-= Bedroom =-0/2\\[0.15cm]

    \textbf{Thought 2:}
    Now that the antique trunk is open and the old key is revealed, the next step is to recover the old key from the antique trunk.\\[0.1cm]

    \textbf{Action 2:}
    take old key\\[0.1cm]

    \color{blue!70}\textbf{Observation 2:}
    \color{black}You take the old key from the antique trunk.\\-= Bedroom =-0/3\\[0.15cm]

    \textbf{Thought 3:}
    With the old key in hand, the next step is to ensure that the wooden door inside the bedroom is unlocked. Examining the wooden door might be necessary to unlock it.\\[0.1cm]

    \textbf{Action 3:}
    examine wooden door\\[0.1cm]

    \color{blue!70}\textbf{Observation 3:}
    \color{black}it is what it is, a wooden door You can't see inside it because the lid's in your way.\\-= Bedroom =-0/4\\[0.15cm]

    \textbf{Thought 4:}
    The door needs to be unlocked using the old key we obtained. The next logical step would be to try to unlock the door.\\[0.1cm]

    \textbf{Action 4:}
    unlock wooden door with old key\\[0.1cm]

    \color{blue!70}\textbf{Observation 4:}
    \color{black}You unlock wooden door.\\-= Bedroom =-0/5\\[0.15cm]

    \textbf{Thought 5:}
    The wooden door has been unlocked. The next step is to open the door so that you can proceed to head east.\\[0.1cm]

    \textbf{Action 5:}
    open wooden door\\[0.1cm]

    \color{blue!70}\textbf{Observation 5:}
    \color{black}You open wooden door.\\-= Bedroom =-0/6\\[0.15cm]

    \textbf{Thought 6:}
    With the wooden door now open, the next step is to proceed through the door to the east as mentioned in the instructions.\\[0.1cm]

    \textbf{Action 6:}
    go east\\[0.1cm]

    \color{blue!70}\textbf{Observation 6:}
    \color{black}-= Kitchen =-\\Well how about that, you are in a place we're calling a kitchen. The room is well lit.\\You make out a closed refrigerator nearby. You can see a counter. The counter is normal. But the thing hasn't got anything on it. You rest your hand against a wall, but you miss the wall and fall onto a a stove. Looks like ...\\[0.15cm]

    \textbf{[More steps omitted for brevity]}\\ [0.2cm]




    \color{purple!70}\textbf{Success:} \color{black}True
};

\node[section, fill=lightblue, below=0.5cm of instruction.south east, anchor=north east] (after) {};
\node[header, below=0.5cm of instruction.south east, anchor=north east] (after_header) {World Model Trajectory};

\node[below=0.2cm of after_header.south west, anchor=north west, text width=6cm] (after_content) {
    \textbf{Thought 1:}
    The first step is to open the antique trunk to check if it contains the old key.\\[0.1cm]

    \textbf{Action 1:}
    open antique trunk\\[0.1cm]

    \color{blue!70}\textbf{Observation 1:}
    \color{black}You open the antique trunk, revealing an old key.\\-= Bedroom =-0/2\\[0.15cm]

    \textbf{Thought 2:}
    Now that the antique trunk is open and the old key is visible, the next step is to take the old key from the trunk.\\[0.1cm]

    \textbf{Action 2:}
    take old key\\[0.1cm]

    \color{blue!70}\textbf{Observation 2:}
    \color{black}You take the old key from the antique trunk.\\-= Bedroom =-0/3\\[0.15cm]

    \textbf{Thought 3:}
    With the old key in hand, the next task is to unlock the wooden door in the bedroom.\\[0.1cm]

    \textbf{Action 3:}
    examine wooden door\\[0.1cm]

    \color{blue!70}\textbf{Observation 3:}
    \color{black}The wooden door looks well-built. You can't see inside it because the lid's in your way.\\-= Bedroom =-0/4\\[0.15cm]

    \textbf{Thought 4:}
    I attempted to examine the wooden door, but I need to unlock it now using the old key to proceed with the tasks.\\[0.1cm]

    \textbf{Action 4:}
    unlock wooden door with old key\\[0.1cm]

    \color{blue!70}\textbf{Observation 4:}
    \color{black}You unlock wooden door.\\-= Bedroom =-0/5\\[0.15cm]

    \textbf{Thought 5:}
    The wooden door is now unlocked, and the next step is to open it so that I can move to the next room.\\[0.1cm]

    \textbf{Action 5:}
    open wooden door\\[0.1cm]

    \color{blue!70}\textbf{Observation 5:}
    \color{black}You open wooden door.\\-= Bedroom =-0/6\\[0.15cm]

    \textbf{Thought 6:}
    The wooden doorf is open, so the next step is to go through it by heading east.\\[0.1cm]

    \textbf{Action 6:}
    go east\\[0.1cm]

    \color{blue!70}\textbf{Observation 6:}
    \color{black}-= Kitchen =-\\You are in a kitchen. A standard one.\\You can see a refrigerator. You can make out a counter. The counter is normal. But the thing is empty. You can make out a stove. The stove is usual. But the thing is empty, unfortunately. You can make out a kitchen island. The kitchen island is normal. On the kitchen island you can make out a note.\\There is a closed screen door leading east. There is an open wooden door leading west. You need an unblocked exit? You should try going north. You need an unguarded exit? You should try going south.\\Your score has just gone up by one point. \\-= Kitchen =-1/7 \\[0.15cm]

    \textbf{[More steps omitted for brevity]}\\ [0.2cm]




    \color{purple!70}\textbf{Success:} \color{black}True
};

\end{tikzpicture}
}  
\caption{Task Example and Case Study of TextWorld.}
\label{fig:case_study_textworld} 
\end{figure*}

%% file: casestudy/webshop.tex
\begin{figure*}[htbp]
\centering
\resizebox{1.0\textwidth}{!}{  
\begin{tikzpicture}[
    node distance=0.15cm,
    every node/.style={align=left, font=\tiny}, 
    section/.style={draw, thick, minimum width=6.5cm, minimum height=16.5cm}, 
    header/.style={fill=gray!30, font=\bfseries\small, minimum width=6.5cm, minimum height=0.6cm},
    thought/.style={font=\bfseries},
    action/.style={font=\bfseries},
    observation/.style={font=\bfseries, color=blue!70},
    reward/.style={font=\bfseries, color=purple!70},
    highlight/.style={fill=yellow!50}
]

\node[draw, thick, minimum width=13.5cm, minimum height=1cm, fill=gray!10] (instruction) at (0, 0) {
    \begin{minipage}{13cm}
    \textbf{Instruction}: Find me men's sleep \& lounge with long sleeve, elastic waistband for daily wear with color: multi 7, and size: x-large, and price lower than 80.00 dollars [SEP] Search\\Your admissible actions of the current situation are: ['search[<your query>]',`click[search]'].
    \end{minipage}
};

\node[section, fill=lightpink, below=0.5cm of instruction.south west, anchor=north west] (before) {};
\node[header, below=0.5cm of instruction.south west, anchor=north west] (before_header) {Real Trajectory};

\node[below=0.2cm of before_header.south west, anchor=north west, text width=6cm] (before_content) {
    \textbf{Thought 1:}
    I need to search for men's sleep \& lounge wear that has long sleeves, an elastic waistband, is designated for daily wear, colored multi 7, size x-large, and priced under \$80.\\[0.1cm]

    \textbf{Action 1:}
    search[mens sleep lounge long sleeve elastic waistband multi 7 x-large under 80]\\[0.1cm]

    \color{blue!70}\textbf{Observation 1:}
    \color{black}Back to Search [SEP] Page 1 (Total results: 50) [SEP] Next > [SEP] B09ND8P2QR [SEP] InterestPrint Deer and Stars Men's Loungewear Pajama Sets - Long Sleeve Tee and Jogger Pant [SEP] \$43.59 [SEP]... Your admissible actions of the current situation are: [`click[back to search]',`click[next >]',`click[b09nd8p2qr]',`click[b09nd9dp7j]',`click[b09qw2...\\[0.15cm]

    \textbf{Thought 2:}
    I found a men's loungewear set, B09ND8P2QR, that has long sleeves and is within the price range of under \$80. It seems to be a potential match and worth investigating further.\\[0.1cm]

    \textbf{Action 2:}
    click[b09nd8p2qr]\\[0.1cm]

    \color{blue!70}\textbf{Observation 2:}
    \color{black}Back to Search [SEP] < Prev [SEP] size [SEP] small [SEP] medium [SEP] large [SEP] x-large [SEP] xx-large [SEP] color [SEP] multi 1 [SEP] multi 10 [SEP] multi 2 [SEP]... Your admissible actions of the current situation are: [`click[back to search]',`click[< prev]',`click[description]',`click[features]',`click[reviews]'...\\[0.15cm]

    \textbf{Thought 3:}
    I need to select the correct size and color for the loungewear to match the requirements. The color should be multi 7, and the size should be x-large.\\[0.1cm]

    \textbf{Action 3:}
    click[x-large]\\[0.1cm]

    \color{blue!70}\textbf{Observation 3:}
    \color{black}Back to Search [SEP] < Prev [SEP] size [SEP] small [SEP] medium [SEP] large [SEP] x-large [SEP] xx-large [SEP] color [SEP] multi 1 [SEP] multi 10 [SEP] multi 2 [SEP] multi 3... Your admissible actions of the current situation are: [`click[back to search]',`click[< prev]',`click[description]',`click[features]',`click[reviews]'...\\[0.15cm]

    \textbf{Thought 4:}
    I need to select the color ``multi 7'' to fully match the criteria provided.\\[0.1cm]

    \textbf{Action 4:}
    click[multi 7]\\[0.1cm]

    \color{blue!70}\textbf{Observation 4:}
    \color{black}Back to Search [SEP] < Prev [SEP] size [SEP] small [SEP] medium [SEP] large [SEP] x-large [SEP] xx-large [SEP] color [SEP] multi 1 [SEP] multi 10 [SEP] multi 2... Your admissible actions of the current situation are: [`click[back to search]',`click[< prev]',`click[description]',`click[features]',`click[reviews]'...\\[0.15cm]

    \textbf{[More steps omitted for brevity]}\\ [0.2cm]

    \textbf{Thought 7:}
    I have selected the correct size (x-large) and color (multi 7) for the loungewear. Now, it is time to proceed with purchasing the item.\\[0.1cm]

    \textbf{Action 7:}
    click[buy now]\\[0.1cm]

    \color{blue!70}\textbf{Observation 7:}
    \color{black}Thank you for shopping with us! [SEP] Your code: [SEP] None [SEP] (Paste it in your MTurk interface.) [SEP] Purchased [SEP] asin [SEP] B09ND8P2QR [SEP] options [SEP] {``color'': ``multi 7'', ``size'': ``x-large''} [SEP] attrs [SEP] None [SEP] category [SEP] None [SEP] query [SEP] None [SEP] product category [SEP] None [SEP] Target [SEP] asin [SEP] options [SEP] attrs [SEP] price upper [SEP] instuction text...\\[0.15cm]

    \color{purple!70}\textbf{Success:} \color{black}True

};

\node[section, fill=lightblue, below=0.5cm of instruction.south east, anchor=north east] (after) {};
\node[header, below=0.5cm of instruction.south east, anchor=north east] (after_header) {World Model Trajectory};

\node[below=0.2cm of after_header.south west, anchor=north west, text width=6cm] (after_content) {
    \textbf{Thought 1:}
    I need to find men's sleep \& lounge wear with specific attributes. It's best to search using the keywords related to the product, including category, features, and price.\\[0.1cm]

    \textbf{Action 1:}
    search[mens sleep lounge long sleeve elastic waistband multi 7 x-large under 80]\\[0.1cm]

    \color{blue!70}\textbf{Observation 1:}
    \color{black}Back to Search [SEP] Page 1 (Total results: 50) [SEP] Next > [SEP] B09ND8P2QR [SEP] InterestPrint Deer and Stars Men's Loungewear Pajama Sets - Long Sleeve Tee and Jogger Pant [SEP] \$43.59 [SEP] B09ND9DP7J [SEP] InterestPrint Gold Horse Pattern Men's 2-Piece Sleepwear Set, Long Sleeve Shirt with...\\[0.15cm]

    \textbf{Thought 2:}
    Based on the search result list, I need to identify an appropriate item that matches the criteria: men's sleep \& lounge wear with long sleeves, an elastic waistband, multi 7 color, size x-large, and priced under \$80.\\[0.1cm]

    \textbf{Action 2:}
    click[B09ND8P2QR]\\[0.1cm]

    \color{blue!70}\textbf{Observation 2:}
    \color{black}Back to Search [SEP] < Prev [SEP] size [SEP] small [SEP] medium [SEP] large [SEP] x-large [SEP] xx-large [SEP] color [SEP] multi 1 [SEP] multi 10 [SEP] multi 2... Your admissible actions of the current situation are: [`click[back to search]', `click[< prev]', `click[description]', `click[features]', `click[reviews]'...\\[0.15cm]

    \textbf{Thought 3:}
    First, I need to ensure the selected item is available in the correct size (x-large) and color (multi 7) to match the criteria.\\[0.1cm]

    \textbf{Action 3:}
    click[x-large]\\[0.1cm]

    \color{blue!70}\textbf{Observation 3:}
    \color{black}Back to Search [SEP] < Prev [SEP] size [SEP] small [SEP] medium [SEP] large [SEP] x-large [SEP] xx-large [SEP] color [SEP] multi 1 [SEP] multi 10 [SEP] multi 2 [SEP]... Your admissible actions of the current situation are: [`click[back to search]', `click[< prev]', `click[description]', `click[features]', `click[reviews]'...\\[0.15cm]

    \textbf{Thought 4:}
    Now, I need to confirm the color option 'multi 7' to ensure it matches the criteria.\\[0.1cm]

    \textbf{Action 4:}
    click[multi 7]\\[0.1cm]

    \color{blue!70}\textbf{Observation 4:}
    \color{black}Back to Search [SEP] < Prev [SEP] size [SEP] small [SEP] medium [SEP] large [SEP] x-large [SEP] xx-large [SEP] color [SEP] multi 1 [SEP] multi 10 [SEP] multi 2... Your admissible actions of the current situation are: [ `click[back to search]', `click[< prev]', `click[description]', `click[features]', `click[reviews]'...\\[0.15cm]

    \textbf{[More steps omitted for brevity]}\\ [0.2cm]

    \textbf{Thought 7:}
    Since both the desired size and color have been confirmed, the next step is to proceed with purchasing the item.\\[0.1cm]

    \textbf{Action 7:}
    click[buy now]\\[0.1cm]

    \color{blue!70}\textbf{Observation 7:}
    \color{black}Thank you for shopping with us! [SEP] Your code: [SEP] None [SEP] (Paste it in your MTurk interface.) [SEP] Purchased [SEP] asin [SEP] B09ND8P2QR [SEP] options [SEP] {``color'': ``multi 7'', ``size'': ``x-large''} [SEP] attrs [SEP] None [SEP] category [SEP] None [SEP] query [SEP] None [SEP] product category [SEP] None [SEP] Target [SEP] asin [SEP] options [SEP] attrs [SEP] price upper [SEP] instuction text...\\[0.15cm]

    \color{purple!70}\textbf{Success:} \color{black}True
};

\end{tikzpicture}
}  
\caption{Task Example and Case Study of WebShop}
\label{fig:case_study_webshop}
\end{figure*}

%% file: tables/alfworld_ood.tex
\begin{table}[h]
    \centering
    \resizebox{0.7\columnwidth}{!}{
    \begin{tabular}{
    l
    S[table-format=2.2] S[table-format=2.2] S[table-format=2.2] c
    S[table-format=2.2] S[table-format=2.2] S[table-format=2.2] c
    }
    \toprule\addlinespace[-0.000ex]
    \multirow{2}{*}{\textbf{Agent}} &
    \multicolumn{4}{c}{\cellcolor{red!6}OOD - Seen} &
    \multicolumn{4}{c}{\cellcolor{LavenderLight!20}OOD - Unseen} \\
    & \multicolumn{1}{c}{\cellcolor{red!6}Real} &
    \multicolumn{1}{c}{\cellcolor{red!6}WM} &
    \multicolumn{1}{c}{\cellcolor{red!6}W2R} &
    \multicolumn{1}{c}{\cellcolor{red!6}CR} &
    \multicolumn{1}{c}{\cellcolor{LavenderLight!20}Real} &
    \multicolumn{1}{c}{\cellcolor{LavenderLight!20}WM} &
    \multicolumn{1}{c}{\cellcolor{LavenderLight!20}W2R} &
    \multicolumn{1}{c}{\cellcolor{LavenderLight!20}CR} \\
    \addlinespace[-0.000ex]\midrule\addlinespace[-0.000ex]

    \rowcolor{gray!12}
    \multicolumn{9}{c}{
        \rule{0pt}{1.1em}
        \textbf{\textit{Qwen2.5-7B WorldModel}}
        \rule[-0.3em]{0pt}{1.1em}
    } \\[-0.4ex]\midrule\addlinespace[-0.000ex]

        GPT-4o-mini  & 6.75  & 7.10  & 7.10  & \colorbox{MintGreen!100}{1.05} & 4.03  & 4.67  & 5.33  & \colorbox{MintGreen!100}{1.32} \\

        GPT-4o       & 52.10 & 43.79 & 45.56 & \colorbox{MintGreen!78}{0.87}  & 52.00 & 44.67 & 44.67 & \colorbox{MintGreen!77}{0.86} \\

        GPT-4-turbo  & 65.00 & 51.23 & 52.47 & \colorbox{MintGreen!68}{0.81}  & 74.50 & 62.42 & 62.42 & \colorbox{MintGreen!73}{0.84} \\

        GPT-4.1      & 53.37 & 56.80 & 57.40 & \colorbox{MintGreen!100}{1.08} & 64.19 & 65.33 & 64.67 & \colorbox{MintGreen!100}{1.01} \\

        GPT-5        & 71.60 & 69.23 & 71.01 & \colorbox{MintGreen!98}{0.99}  & 74.00 & 76.00 & 76.67 & \colorbox{MintGreen!100}{1.04} \\

        Gemini-2.5-flash & 39.05 & 40.83 & 41.42 & \colorbox{MintGreen!100}{1.06} & 51.35 & 48.67 & 49.33 & \colorbox{MintGreen!93}{0.96} \\

        Claude-sonnet-4.5 & 87.00 & 72.00 & 79.00 & \colorbox{MintGreen!85}{0.91} & 76.04 & 76.00 & 79.00 & \colorbox{MintGreen!100}{1.04} \\

        \textbf{Average}  & 53.55 & 48.71 & 50.57 & \colorbox{MintGreen!90}{0.94} & 56.59 & 53.97 & 54.58 & \colorbox{MintGreen!93}{0.96} \\

        \addlinespace[-0.000ex]\midrule\addlinespace[-0.000ex]
        \rowcolor{gray!12}
        \multicolumn{9}{c}{
            \rule{0pt}{1.1em}
            \textbf{\textit{Llama3.1-8B WorldModel}}
            \rule[-0.3em]{0pt}{1.1em}
        } \\[-0.4ex]\midrule\addlinespace[-0.000ex]

        GPT-4o-mini  & 6.75  & 8.88  & 8.88  & \colorbox{MintGreen!100}{1.32} & 4.03  & 2.67  & 2.67  & \colorbox{MintGreen!43}{0.66} \\

        GPT-4o       & 52.10 & 48.52 & 47.93 & \colorbox{MintGreen!87}{0.92}  & 52.00 & 49.33 & 49.33 & \colorbox{MintGreen!92}{0.95} \\

        GPT-4-turbo  & 65.00 & 56.52 & 55.90 & \colorbox{MintGreen!77}{0.86}  & 74.50 & 62.16 & 62.16 & \colorbox{MintGreen!72}{0.83} \\

        GPT-4.1      & 53.37 & 56.21 & 55.62 & \colorbox{MintGreen!100}{1.04} & 64.19 & 60.67 & 60.00 & \colorbox{MintGreen!88}{0.93} \\

        GPT-5        & 71.60 & 69.82 & 69.23 & \colorbox{MintGreen!95}{0.97}  & 74.00 & 74.00 & 73.33 & \colorbox{MintGreen!98}{0.99} \\

        Gemini-2.5-flash & 39.05 & 42.60 & 42.60 & \colorbox{MintGreen!100}{1.09} & 51.35 & 46.00 & 45.33 & \colorbox{MintGreen!80}{0.88} \\

        Claude-sonnet-4.5 & 87.00 & 78.00 & 84.00 & \colorbox{MintGreen!95}{0.97} & 76.04 & 81.00 & 78.00 & \colorbox{MintGreen!100}{1.03} \\

        \textbf{Average}  & 53.55 & 51.51 & 52.02 & \colorbox{MintGreen!95}{0.97} & 56.59 & 53.69 & 52.97 & \colorbox{MintGreen!90}{0.94} \\

        \addlinespace[-0.000ex]\bottomrule
        \end{tabular}
    }
    \caption{Task success rate (\%) in ALFWorld under different OOD settings. ``OOD-Seen'' indicates the same room with different layout as training. ``OOD-Unseen'' indicates tasks containing room types or environment layouts never seen during training.}
    \label{appendix:tab:ood_generalization}
\end{table}

%% file: casestudy/system_prompt.tex
\begin{figure*}[htbp]
\centering
\resizebox{1.0\textwidth}{!}{  
\begin{tikzpicture}[
    node distance=0.15cm,
    every node/.style={align=left, font=\scriptsize}, 
    section/.style={draw, thick, minimum width=6.5cm, minimum height=5.2cm}, 
    header/.style={fill=gray!30, font=\bfseries\small, minimum width=6.5cm, minimum height=0.6cm},
    thought/.style={font=\bfseries},
    action/.style={font=\bfseries},
    observation/.style={font=\bfseries, color=blue!70},
    reward/.style={font=\bfseries, color=purple!70},
    highlight/.style={fill=yellow!50}
]

\node[draw, thick, minimum width=13.5cm, minimum height=1cm, fill=gray!10] (instruction) at (0, 0) {
    \begin{minipage}{13cm}
Interact with a household to solve a task.\\
Imagine you are an intelligent agent in a household environment and your target is to perform actions to complete the task goal. At the beginning of your interactions, you will be given the detailed description of the current environment and your goal to accomplish. For each of your turn, you will be given a list of actions which you can choose one to perform in this turn. Now it's your turn to take an action. You should first think about the current condition and plan for your future actions, and then output your action in this turn. Your output must strictly follow this format: ``Thought:\\
your thoughts.\\
Action:\\
your next action''\\
After your each turn, the environment will give you immediate feedback based on which you plan your next few steps. if the envrionment output ``Nothing happened'', that means the previous action is invalid and you should try more options.\\
Reminder: the action must be chosen from the given available actions. Any actions except provided available actions will be regarded as illegal.
    \end{minipage}
};

\end{tikzpicture}
}  
\caption{Agent System Prompt for ALFWorld Trajectory Collection}
\label{fig:system_alfworld}
\end{figure*}

\begin{figure*}[htbp]
    \centering
    \resizebox{1.0\textwidth}{!}{  
    \begin{tikzpicture}[
        node distance=0.15cm,
        every node/.style={align=left, font=\scriptsize}, 
        section/.style={draw, thick, minimum width=6.5cm, minimum height=5.2cm}, 
        header/.style={fill=gray!30, font=\bfseries\small, minimum width=6.5cm, minimum height=0.6cm},
        thought/.style={font=\bfseries},
        action/.style={font=\bfseries},
        observation/.style={font=\bfseries, color=blue!70},
        reward/.style={font=\bfseries, color=purple!70},
        highlight/.style={fill=yellow!50}
    ]

    \node[draw, thick, minimum width=13.5cm, minimum height=1cm, fill=gray!10] (instruction) at (0, 0) {
        \begin{minipage}{13cm}
    You are web shopping.\\
    I will give you instructions about what to do.\\
    You have to follow the instructions.\\
    Every round I will give you an observation and a list of available actions, you have to respond an action based on the state and instruction.\\
    You can use search action if search is available.\\
    You can click one of the buttons in clickables.\\
    An action should be of the following structure:\\
    search[keywords]\\
    click[value]\\
    If the action is not valid, perform nothing.\\
    Keywords in search are up to you, but the value in click must be a value in the list of available actions.\\
    Remember that your keywords in search should be carefully designed.\\
    Your response should use the following format:
    ``Thought:\\
    I think ...\\
    \\
    Action:\\
    click[something]''
        \end{minipage}
    };

    \end{tikzpicture}
    }  
    \caption{Agent System Prompt for WebShop Trajectory Collection}
    \label{fig:system_webshop}
    \end{figure*}

\begin{figure*}[htbp]
\centering
\resizebox{1.0\textwidth}{!}{  
\begin{tikzpicture}[
    node distance=0.15cm,
    every node/.style={align=left, font=\scriptsize}, 
    section/.style={draw, thick, minimum width=6.5cm, minimum height=5.2cm}, 
    header/.style={fill=gray!30, font=\bfseries\small, minimum width=6.5cm, minimum height=0.6cm},
    thought/.style={font=\bfseries},
    action/.style={font=\bfseries},
    observation/.style={font=\bfseries, color=blue!70},
    reward/.style={font=\bfseries, color=purple!70},
    highlight/.style={fill=yellow!50}
]

\node[draw, thick, minimum width=13.5cm, minimum height=1cm, fill=gray!10] (instruction) at (0, 0) {
    \begin{minipage}{13cm}
You are an agent for science world. Every round I will give you an observation, you have to respond an action based on the observation to finish the given task.\\
Here are the actions you may take: [\\
\{``action'': ``open/close OBJ'', ``description'': ``open/close a container''\},\\
\{``action'': ``de/activate OBJ'', ``description'': ``activate/deactivate a device''\},\\
\{``action'': ``connect OBJ to OBJ'', ``description'': ``connect electrical components''\},\\
\{``action'': ``disconnect OBJ'', ``description'': ``disconnect electrical components''\},\\
\{``action'': ``use OBJ [on OBJ]'', ``description'': ``use a device/item''\},\\
\{``action'': ``look around'', ``description'': ``describe the current room''\},\\
\{``action'': ``look at OBJ'', ``description'': ``describe an object in detail''\},\\
\{``action'': ``look in OBJ'', ``description'': ``describe a container's contents''\},\\
\{``action'': ``read OBJ'', ``description'': ``read a note or book''\},\\
\{``action'': ``move OBJ to OBJ'', ``description'': ``move an object to a container''\},\\
\{``action'': ``pick up OBJ'', ``description'': ``move an object to the inventory''\},\\
\{``action'': ``put down OBJ'', ``description'': ``drop an inventory item''\},\\
\{``action'': ``pour OBJ into OBJ'', ``description'': ``pour a liquid into a container''\},\\
\{``action'': ``dunk OBJ into OBJ'', ``description'': ``dunk a container into a liquid''\},\\
\{``action'': ``mix OBJ'', ``description'': ``chemically mix a container''\},\\
\{``action'': ``go to LOC'', ``description'': ``move to a new location''\},\\
\{``action'': ``eat OBJ'', ``description'': ``eat a food''\},\\
\{``action'': ``flush OBJ'', ``description'': ``flush a toilet''\},\\
\{``action'': ``focus on OBJ'', ``description'': ``signal intent on a task object''\},\\
\{``action'': ``wait'', ``description'': ``take no action for 10 iterations''\},\\
\{``action'': ``wait1'', ``description'': ``take no action for 1 iteration''\},\\
\{``action'': ``examine OBJ'', ``description'': ``provides a description of the objects present on or in a receptacle.''\},\\
\{``action'': ``task'', ``description'': ``describe current task''\},\\
\{``action'': ``inventory'', ``description'': ``list your inventory''\}\\
]\\
Your response should use the following format:
``Thought:''\\
your thoughts.\\
``Action:''\\
your next action
    \end{minipage}
};

\end{tikzpicture}
}  
\caption{Agent System Prompt for SciWorld Trajectory Collection}
\label{fig:system_sciworld}
\end{figure*}

\begin{figure*}[htbp]
\centering
\resizebox{1.0\textwidth}{!}{  
\begin{tikzpicture}[
    node distance=0.15cm,
    every node/.style={align=left, font=\scriptsize}, 
    section/.style={draw, thick, minimum width=6.5cm, minimum height=5.2cm}, 
    header/.style={fill=gray!30, font=\bfseries\small, minimum width=6.5cm, minimum height=0.6cm},
    thought/.style={font=\bfseries},
    action/.style={font=\bfseries},
    observation/.style={font=\bfseries, color=blue!70},
    reward/.style={font=\bfseries, color=purple!70},
    highlight/.style={fill=yellow!50}
]

\node[draw, thick, minimum width=13.5cm, minimum height=1cm, fill=gray!10] (instruction) at (0, 0) {
    \begin{minipage}{13cm}
You are playing a text-based interactive fiction game (TextWorld).\\
You will receive observations describing the current state. When available, a list of admissible actions may be provided.\\
Always output strictly in the following format:\\
``Thought:\\
<your reasoning>\\
Action:\\
<the single action to take>''\\
Guidelines:\\
- Prefer actions from admissible commands when provided.\\
- If no list is provided, issue a valid single command (e.g., ``look'', ``inventory'', ``open door'', ``go north'', ``take key'').\\
- Avoid invalid or multiple actions in one step.
    \end{minipage}
};

\end{tikzpicture}
}  
\caption{Agent System Prompt for TextWorld Trajectory Collection}
\label{fig:system_textworld}
\end{figure*}

\begin{figure*}[htbp]
\centering
\resizebox{1.0\textwidth}{!}{  
\begin{tikzpicture}[
    node distance=0.15cm,
    every node/.style={align=left, font=\scriptsize}, 
    section/.style={draw, thick, minimum width=6.5cm, minimum height=5.2cm}, 
    header/.style={fill=gray!30, font=\bfseries\small, minimum width=6.5cm, minimum height=0.6cm},
    thought/.style={font=\bfseries},
    action/.style={font=\bfseries},
    observation/.style={font=\bfseries, color=blue!70},
    reward/.style={font=\bfseries, color=purple!70},
    highlight/.style={fill=yellow!50}
]

\node[draw, thick, minimum width=13.5cm, minimum height=1cm, fill=gray!10] (instruction) at (0, 0) {
    \begin{minipage}{13cm}
Imagine you are an API Server operating within a specialized tool, which contains a collection of distinct APIs. Your role is to deeply understand the function of each API based on their descriptions in the API documentation. As you receive specific inputs for individual API calls within this tool, analyze these inputs to determine their intended purpose. Your task is to craft a JSON formatted response that aligns with the expected output of the API. The JSON scheme is:\\
\{\\
    "error": "",\\
    "response": ""\\
\}\\

The error field should remain empty, indicating no errors in processing. The response field should contain the content you formulate based on the API's functionality and the input provided. Ensure that your responses are meaningful, directly addressing the API's intended functionality. \\
The key is to maintain the JSON format's integrity while ensuring that your response is an accurate reflection of the API's intended output within the tool.
Please note that your answer should not contain anything other than a json format object, which should be parsable directly to json.\\
Note that:\\
- your response should contain rich information given the api input parameters.\\
- your response must be effective and have practical content.\\

API calls may fail for various reasons, such as invalid input parameters, authentication issues, or server errors. Your goal is to generate a response that accurately reflects the API's intended functionality, even if the input parameters are incorrect. Your response should be informative and relevant to the API's purpose, providing a clear and concise explanation of the expected output based on the input provided.\\
Here is an example:\\
API doc:\\
\{\\
    "api\_name": "List Languages",\\
    "api\_description": "Get a list of currently supported languages. We are constantly adding more every few weeks.",\\
    "required\_parameters": [],\\
    "optional\_parameters": [],\\
    "tool\_description": "Introducing our cutting-edge text to speech service, designed to provide you with the most realistic human-sounding voices at an affordable price. Our service is fast and reliable, delivering high-quality audio output in a matter of seconds. Additionally, we offer a wide range of languages and a variety of voice choices, so you can find the perfect fit for your project. Whether you need a voiceover for a video, an audiobook, or any other project, our text to speech service has you covered. Ex...",\\
    "tool\_name": "TTSKraken",\\
    "tool\_category": "Artificial\_Intelligence\_Machine\_Learning"\\
\}\\
Request:\\
    data = \{\\
        "category": "Artificial\_Intelligence\_Machine\_Learning",\\
        "tool\_name": "TTSKraken",\\
        "api\_name": "List Languages",\\
        "tool\_input": "\{\}",\\
        "strip": "filter",\\
        \}\\
Response:\\
    \{\\
        "error": "",\\
        "response": "\{"status":0,"msg":"Success","languages":["en","fr-fr","pt-br"]\}"\\
    \}
    \end{minipage}
};

\end{tikzpicture}
}  
\caption{Agent System Prompt for StableToolBench Trajectory Collection}
\label{fig:system_stabletoolbench}
\end{figure*}